%% file: main.tex
\documentclass{article}
\PassOptionsToPackage{numbers, compress}{natbib}

     \usepackage[preprint]{neurips_2025}

\usepackage[utf8]{inputenc} % allow utf-8 input
\usepackage[T1]{fontenc}    % use 8-bit T1 fonts
\usepackage{hyperref}       % hyperlinks
\usepackage{url}            % simple URL typesetting
\usepackage{booktabs}       % professional-quality tables
\usepackage{amsfonts}       % blackboard math symbols
\usepackage{nicefrac}       % compact symbols for 1/2, etc.
\usepackage{microtype}      % microtypography
\usepackage[dvipsnames]{xcolor}         % colors
\usepackage{colortbl}
\usepackage{dsfont}
\usepackage{stmaryrd}       % additional mathematical symbols

\usepackage{url}
\usepackage{graphicx} % Required for inserting images
\usepackage{booktabs}
\usepackage{researchpack}
\usepackage{amsfonts}
\usepackage{listings}
\usepackage{xcolor}
\usepackage{hyperref}
\hypersetup{
   colorlinks=true,
   linkcolor=blue,
   citecolor=blue,
   urlcolor=blue
}
\usepackage[capitalize, nameinlink]{cleveref}
\definecolor{codegreen}{rgb}{0,0.6,0}
\definecolor{codegray}{rgb}{0.5,0.5,0.5}
\definecolor{codepurple}{rgb}{0.58,0,0.82}
\definecolor{backcolour}{rgb}{0.95,0.95,0.92}

\lstdefinestyle{mystyle}{
    backgroundcolor=\color{backcolour},   
    commentstyle=\color{codegray},
    keywordstyle=\color{codegray},
    numberstyle=\tiny\color{codegray},
    stringstyle=\color{codegray},
    basicstyle=\ttfamily\footnotesize,
    breakatwhitespace=false,         
    breaklines=true,                 
    captionpos=b,                    
    keepspaces=true,                 
    numbers=left,                    
    numbersep=5pt,                  
    showspaces=false,                
    showstringspaces=false,
    showtabs=false,                  
    tabsize=2
}

\usepackage{researchpack}
\hypersetup{
    colorlinks=true,
    linkcolor=blue,
    citecolor=blue,
    urlcolor=blue
}

\newcommand{\LtX}{LtX\xspace}

\newcommand{\method}{{\tt REX}\xspace}
\newcommand{\methodQ}{{REX}}
\newcommand{\acronym}{REjector of low-quality Explanations\xspace}

\newcommand{\InstNov}{{\tt InstNov}}
\newcommand{\ExplNov}{{\tt ExplNov}}
\newcommand{\PredAmb}{{\tt PredAmb}\xspace}
\newcommand{\PASTARej}{{\tt PASTARej}\xspace}
\newcommand{\PASTA}{{\tt PASTA}\xspace}
\newcommand{\SVM}{{\tt SVM}\xspace}

\newcommand{\LIME}{{\tt LIME}\xspace}
\newcommand{\SHAP}{{\tt SHAP}\xspace}

\title{Knowing What You Cannot Explain: \\
Learning to Reject Low-Quality Explanations}
\author{%
  Luca Stradiotti \\
  DTAI lab \& Leuven.AI, \\
  KU Leuven, Belgium \\
  \texttt{luca.stradiotti@kuleuven.be}\\
\And
  Dario Pesenti \\
  CIMeC, \\
  University of Trento, Italy\\
  \texttt{dario.pesenti@unitn.it} \\
\AND
  Stefano Teso \\
  CIMeC and DISI, \\
  University of Trento, Italy\\
  \texttt{stefano.teso@unitn.it} \\
\And
  Jesse Davis \\
  DTAI lab \& Leuven.AI, \\
  KU Leuven, Belgium \\
  \texttt{jesse.davis@kuleuven.be} \\
}

\begin{document}
\maketitle

\begin{abstract}
     Learning to Reject (LtR) frameworks allow ML models to abstain from uncertain predictions and promote user trust. However, since current LtR strategies focus solely on predictive performance, they completely neglect explanation quality. Low-quality explanations -- whether they inaccurately reflect the model's reasoning or fail to satisfy users -- can severely compromise trust assessments and induce  over-reliance on incorrect predictions. We argue that models should abstain from making a prediction when they cannot offer a satisfactory explanation for it and introduce a framework for \textit{learning to reject low-quality explanations} (\LtX) in which predictors are equipped with a \textit{rejector} that evaluates the explanation quality.  Focusing on popular attribution techniques, we propose \method (\acronym), which learns a rejector from explanation quality labels combining machine-side judgments with explicit human annotations to assess explanation quality. Our empirical evaluation demonstrates that \method outperforms popular LtR strategies and baselines relying on isolated explanation metrics. Finally, to support future research, we publicly release a novel, larger-scale dataset of 1050 human-annotated machine explanations.
\end{abstract}

\input{texFiles/1_Introduction}
\input{texFiles/2_Preliminaries}
\input{texFiles/3_Methodology}
\input{texFiles/4_Experiments}
\input{texFiles/5_RelatedWork}
\input{texFiles/6_Conclusion}
\begin{ack}
This research was supported by the Flemish Government through the “Onderzoeksprogramma Artificiële Intelligentie (AI) Vlaanderen” programme [LS, JD], and by the KU Leuven Research Fund (iBOF/21/075) [JD]. This project has received funding from the European Union under Grant Agreement No. 101120763 – TANGO [ST, DP]. Views and opinions expressed are however those of the author(s) only and do not necessarily reflect those of the European Union or the European Health and Digital Executive Agency (HaDEA). Neither the European Union nor the granting authority can be held responsible for them.
 \end{ack}

\bibliography{references, explanatory-supervision}

\appendix
\input{texFiles/Appendix}
\end{document}

%% file: texFiles/1_Introduction.tex
\section{Introduction}\label{sec:introduction}
\textit{Learning to Reject} (LtR) is a framework that allows models to abstain from making a prediction~\cite{chow1970optimum}. Effectively, this framework imbues the model with the option to say ``I don't know'' and defer a decision to, \eg a human expert or a more complex model~\cite{hendrickx2024machine}. Current approaches abstain when the model is at a heightened risk of making a misprediction, which is motivated by two inter-related aims. First, by abstaining on ``difficult'' examples, the model's predictive performance improves when considering only those examples for which it does offer a prediction~\cite{geifman2017selective}.  Second, abstention promotes trust as the model makes fewer mistakes and shows awareness of when its prediction may be wrong~\cite{pugnana2024deep}. Establishing trust is particularly important in high stakes applications where incorrect predictions can have serious consequences~\cite{rajkomar2019machine}. 

\begin{figure}[!t]
    \centering
    \includegraphics[width=1\linewidth]{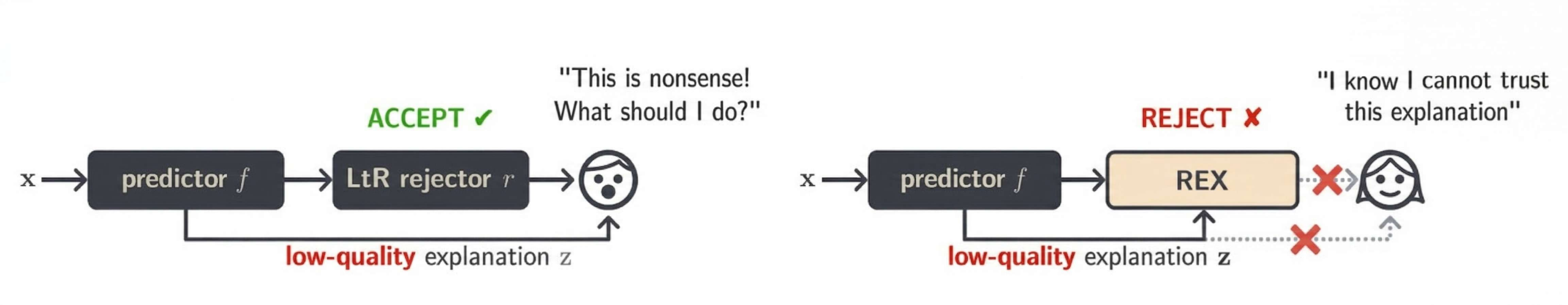}
    \caption{\textbf{Illustration of \method}.
    LtR is unconcerned with the quality of machine explanations (left).
    \textit{\method instead addresses LtX}, which requires to reject predictions that cannot be explained properly to stakeholders, improving trust assessment and down-stream decision quality (right).
    }
    \label{fig:page2}
\end{figure}

Predictive accuracy is not the only aspect of model performance that affects trust. In many settings, a model must be able to provide an explanation for its prediction as required by legal or regulatory frameworks~\cite{GDPR2016,phillips2021four}.
However, not all explanations are high-quality. Explanation quality is multi-faceted: good explanations must both (a) accurately and consistently capture the rationale behind the model's decisions while (b) also being plausible to the user. Unfortunately, low-quality explanations can adversely affect trust assessment and downstream decisions~\cite{gilpin2018explaining,schneider2023deceptive,lakkaraju2020how}. Moreover, they may induce over-reliance by persuading users to accept incorrect predictions~\cite{joshi2023machine,si2024large,sieker2024illusion}.

Despite its importance, current approaches to LtR neglect explanation quality altogether~\cite{kim2024human}. At a high-level, LtR algorithms abstain when they are uncertain about their prediction or a test example differs substantially from the observed training data~\cite{ruggieri2025things}. However, there is no guarantee that these are the precise situations where the model is also more likely to offer a low-quality explanation. For instance, models affected by shortcuts can output accurate, confident predictions that rely on incorrect features~\cite{geirhos2020shortcut}. 

We address this gap by formalizing the \textit{\textbf{Learning to Reject Low-Quality Explanations}} (\LtX) problem where a model abstains from making a prediction when it cannot offer a satisfactory explanation, cf. \cref{fig:page2} (right). We propose \method (\acronym) an algorithm for training a rejector that can identify two distinct failure cases for explanations. On the one hand, ``machine-side'' failures occur when an explanation does not effectively reflect the model's reasoning. On the other hand, ``human-side'' failures arise when users find an explanation unsatisfactory or misaligned with their thought process. Consequently, it can capture the multifactorial nature of explanation quality. \method uses both information about the explanation (\eg machine-side metrics) and explicit user feedback to train the rejector because while machine-side failures can be detected in an unsupervised fashion, human-side failures cannot. Empirically, we verify our approach on both benchmark datasets and on a new larger-scale dataset of human-annotated machine explanations that we collected, which will made publicly available upon acceptance. \method outperforms popular LtR strategies, as well as rejection baselines that rely solely on a single notion of explanation quality. 

\textbf{Contributions}:  Summarizing, we:
    (\textit{i}) Introduce the problem of \textit{learning to reject low-quality explanations} (\LtX), filling a significant gap in current LtR strategies, which ignore explanation quality altogether.
    (\textit{ii}) Design \method, a rejector that uses modest amounts of explanations quality labels combining machine-side judgments with human annotations to learn an effective rejection policy.
    (\textit{iii}) Empirically evaluate \method on both popular datasets and on a novel human-annotated task collected specifically for this work, showcasing its benefits over standard LtR and isolated explanation metrics.
    (\textit{iv}) Provide the first larger-scale ($1050$ examples, $5$ annotations each) dataset of human-annotated explanations  as well as a template for running the associated collection campaign.

%% file: texFiles/2_Preliminaries.tex
\section{Preliminaries}\label{sec:preliminaries}
We consider a \textit{predictor} $f$ that maps examples $\vx \in \calX$ to a target value $f(\vx) \in \calY$. Here, $\calX$ is a $d$-dimensional feature space and $\calY$ a discrete $\left(\calY = \{1, \ldots, C\}\right)$ or continuous $\left(\calY = \bbR\right)$ target space. When the target is discrete, we view the predictor as a probabilistic \textit{classifier} that assigns a distribution $P(Y | X=\vx)$ to each example $\vx$; predictions are obtained via MAP inference, that is $ f(\vx) = \argmax_{c\in\calY} P(Y=c | \vx)$. When the target is continuous, we view it as a \textit{regressor} $f(\vx)=\mathbb{E}[Y | X=\vx]$.

We assume the predictor is paired with an \textit{explainer} $e$ which produces a local explanation $\vz = e(f,\vx)$ for an individual prediction $f(\vx)$. We focus on \textit{feature importance} explanations, which are perhaps the most well-known and widespread class of explanations~\cite{ribeiro2016should,lundberg2017unified}. These associate a \textit{relevance score} $z_i \in \bbR$ to each example feature $x_i$ that quantifies its relative contribution to the prediction.  For example, in loan approval, $\vz$ might indicate that an application $\vx$ was rejected (\ie $f(\vx) = 0$) because the feature $x_{\tt income}$ is too low by having its corresponding explanation score $z_{\tt income}$ ``vote'' against approval. We refer to the pair $\left(f(\vx), \vz\right)$ as the model \textit{output}, since each prediction $f(\vx)$ is returned to the user along with its corresponding explanation $\vz$.

\paragraph{Learning to reject} To promote trust, a \textit{\textbf{Learning to Reject}} (LtR) model combines a predictor $f$ with a \textit{rejector} $r$. The role of the rejector is to offload difficult predictions to a human expert~\cite{hendrickx2024machine}.  Formally, it does so by extending the target space $\calY$ to include an additional symbol $\circledR$ indicating the model abstains from making a prediction~\cite{destefano2000to}. Two classes of rejection strategies have been studied in the literature. \textit{\textbf{Ambiguity rejection}} occurs when the predictor $f$ is too uncertain about a particular example $\vx$, \eg due to class overlap or poor choice of the predictor’s hypothesis space~\cite{pugnana2023auc}. \textit{\textbf{Novelty rejection}} checks if $\vx$ falls in a region where there is little or no training data~\cite{van2021reject}. Although existing rejection strategies improve the model's reliability~\cite{geifman2017selective}, they focus solely on predictor's performance~\cite{hendrickx2024machine} and completely neglect explanation quality.

\paragraph{Metrics of Explanation Quality}  Since explanation quality admits multiple interpretations, numerous metrics have been proposed to evaluate it~\cite{chen2022makes}, cf. \cref{app:explanation_metrics} for further details. Machine-side metrics depend solely on the relationship between the explanation and the predictor and, as such, can be computed accurately using information gathered during inference and/or training. E.g., \textit{\textbf{faithfulness}}~\cite{ramaravind2021towards,azzolin2025reconsidering} measures whether an explanation reflects the model’s reasoning process, and can be estimated by assessing whether the features with high relevance are sufficient and necessary for the prediction. Another key metric is \textit{\textbf{stability}}~\cite{slack2021reliable}, which measures the degree to which different (possibly conflicting) explanations can be provided for a given prediction. Conversely, human-side metrics gauge whether an explanation is actually satisfactory \textit{to the user}, and as such cannot be estimated in a fully unsupervised manner~\cite{naveed2024overview}. A recent example is PASTA, a novel perceptual quality metric that learns to mimic human preferences from feedback~\cite{kazmierczak2024benchmarking} and that we compare against in our experiments (\cref{sec:experiments}). Although several metrics of explanation quality exist, none have been integrated into rejection strategies to guide the rejector's decisions. 

%% file: texFiles/3_Methodology.tex
\section{Learning to Reject Low-Quality Explanations}
\label{sec:methodology}
We now formalize the \textit{Learning to Reject Low-Quality Explanations} (\LtX) problem. Here, the model evaluates the quality of an explanation for a test example and only offers an output if the explanation is of a sufficiently high-quality. Formally, we define a model with reject option in the \LtX setting as follows.
\begin{definition}
An \LtX model $m$ consists of a predictor $f$, an explainer $e$ and a rejector $r$. Given an example $\vx$, the output of $m$ is:
    $$
        m_{(f,e,r)}(\vx) =
        \begin{cases}
            \circledR & \text{ if } \; \phi\left(\vx, f, e\right) < \tau  \\
            \left(f(\vx), e\left(f,\vx\right)\right) & \text{ otherwise }
        \end{cases}
    $$  
\end{definition}
In this setting, the rejector $r$ acts as a filter based on the explanation quality: if the score $\phi\left(\vx, f, e\right)$ it assigns to the quality of the explanation is below a threshold $\tau$, the model abstains from providing the prediction (and explanation) to the user.

Fundamentally, the key challenge is to learn a rejector that assesses explanation quality. To this end, we propose an algorithm called \method (\acronym), which has two key components: (1) constructing an annotated dataset about explanation quality and (2) training the rejector $r$ itself.  We now describe each step in more detail. 

\subsection{Constructing an annotated dataset for \LtX}\label{sec:creating_dataset} 
Because explanation quality is multi-faceted, an ideal rejector must evaluate it across two complementary dimensions: the \textit{machine-side}, whether the explanation effectively captures the model's reasoning, and the \textit{human-side}, whether the explanation is satisfactory to the user. While machine-side metrics -- such as faithfulness and stability -- can be readily estimated procedurally~\cite{azzolin2025reconsidering,slack2021reliable}, human-side aspects of explanation quality depend on human judgment, and as such can only be estimated by leveraging actual expert feedback~\cite{kazmierczak2024benchmarking}.  For this reason, \method relies on a modicum of expertly annotated explanations. 

Therefore, we construct the training set as:
$$
\calD = \{ ((\vz_i,\eta_{\vz_i}), y_{\vz_i}): i \in \{1, \dots, n\}\}
$$ 
where each explanation $\vz_i$ is accompanied by machine-side quality measurements $\eta_{\vz_i}$ (\eg their estimated faithfulness and stability) and quality labels $y_{\vz_i}$. Intuitively, the feature relevance scores serve to help discriminate between plausible vs. implausible explanations, while the values of the machine-side metrics allow identifying unfaithful and unstable explanations. In principle, any machine-side can be included as a feature. 

The labels $y_{\vz_i} \in \{0,1\}$ are meant to capture the overall quality of the explanation, where $0$ denotes low-quality and $1$ high-quality. For instance, an expert could deem an explanation implausible -- and therefore low quality -- if, according to them, it does not correctly discriminate between truly relevant and irrelevant features. While we expect labels in most practical applications to reflect a combination of both machine-side and human-side aspects, \method is designed to work with any notion of explanation quality. In our experiments, we compute the labels $y_{\vz}$ by conjoining labels derived from machine-side metrics with (simulated) expert quality judgments (see \cref{sec:Q1-Q2}).

An optional step to reduce the annotation burden is to perform data augmentation  by leveraging \textit{\textbf{per-feature human labels}}. Specifically, users can optionally flag specific features whose relevance scores they deem incorrect. The cost of providing this additional feedback is  negligible since users are already considering the relevance of the features when forming their overall quality judgment~\cite{teso2019explanatory,raghavan2006active}. To reduce cognitive load in high-dimensional domains, this step can be simplified by displaying only a few top-ranked features (\ie those with the highest relevance scores)~\cite{kulesza2015principles}.\footnote{In this case, users flag either \textit{(i)} presented features whose score they believe to be incorrect, or \textit{(ii)} unshown features they expect to be relevant.} We exploit the per-feature labels to construct an augmented dataset $\calD_{aug}$. The core intuition of our strategy is that slightly perturbing the incorrect relevance scores for high-quality explanations, and the \textit{correct} entries of low-quality ones will not affect the explanation's quality labels while yielding more information about which features are (not) important. For each explanation $\vz$ with flagged incorrect scores, we create $K$ new explanations $\vz_{aug}$ with the same quality label $y_{\vz}$ and metric values $\eta_{\vz}$ but where:
$$
   \vz_{aug}  \sim \calN(\vz, \epsilon_0 \vs \times  \Sigma).
$$
\noindent where $\epsilon_0$ controls the magnitude of the perturbations, $\Sigma$ is a diagonal matrix containing per-feature standard deviations across $\calD$ and is responsible for rescaling perturbations according to the data distribution, and $\vs$ is a binary mask that ensures noise is applied only to incorrect (correct) features for high-quality (low-quality) explanations.

\subsection{Learning the rejector}\label{sec:learning_rejector}
The rejector $r$ consists of a binary classifier $\phi$ and a threshold $\tau$. Specifically, \method trains $\phi$ to estimate the overall explanation quality from  $\calD$ (or $\calD_{aug}$). \method is agnostic to the specific choice of classifier provided it associates a score with its prediction. Empirically, we find that simple models (\eg kernel SVMs~\cite{cortes1995support}) work well.

The rejection threshold $\tau$ determines how often a prediction and explanation are offered by $m$. Lower values of $\tau$ mean that $m$ will operate more autonomously (\ie return more prediction-explanation pairs) albeit with the risk that some explanations are low quality. Higher values mean the model is more cautious and only offers prediction-explanation pairs when its more certain about the quality of the explanation but at the cost of offloading more decisions to the user. Hence, this value should be carefully tuned, \eg on validation to navigate this tradeoff. Two natural strategies are to set $\tau$ such that (i) it achieves a specific rejection rate on the validation data (\eg one aligned with a user's capacity to make decisions) or (ii) its rejection rate is equal to the proportion of low-quality explanations in the training set.  

%% file: texFiles/4_Experiments.tex
\section{Empirical Evaluation}\label{sec:experiments}
Empirically, we address the following research questions: 
\begin{description}
    \item[\textbf{Q1}] Is \method effective at rejecting low-quality explanations?
    \item[\textbf{Q2}] Can \method handle both machine- and human-side quality annotations?
    \item[\textbf{Q3}] Can \method recover actual human quality judgments?
\end{description}
The Appendix examines two additional questions: \cref{app:input_space} explores the effect of what information \method's rejector has access to on its ability to reject low-quality explanations and \cref{app:uler-aug} investigates the effect of the data augmentation based on per-feature feedback on its performance. 

\textbf{Competitors}. We compare \method against four baseline strategies. To demonstrate that classic LtR approaches are unsuited for rejecting low-quality explanations, we consider two widely used LtR strategies. \underline{\InstNov} rejects examples based on their novelty~\cite{sun2022out}: it first computes their Euclidean distances to the $k$ nearest training examples and converts these into scores using a monotonically decreasing function (\eg $1/(1+x)$), such that farthest examples get lower scores. \underline{\PredAmb} uses prediction's confidence as score~\cite{hendrickx2024machine}. For binary classification tasks, confidence is computed as the margin of the class probabilities $|P(Y=1 | \vx)- P(Y=0 | \vx)|$~\cite{perini2023unsupervised}. 

We consider two baselines that target explanations. \underline{\ExplNov} mirrors \InstNov but works in explanation space. It operates on the assumption that explanations that are highly dissimilar to ones observed on the training set are likely to be low-quality. {\tt \underline{PASTARe}j} is an adapted version of the state-of-the-art human-side \PASTA-metric for the plausibility of an explanation ~\cite{kazmierczak2024benchmarking}. Since our focus is on tabular data, we drop the embedding network and fit only the scoring network using the explanations as input when fitting human judgments.

\textbf{Evaluation metrics}. Ideally, a user wants to receive only predictions accompanied by high-quality explanations. A good rejector should therefore minimize the number of low-quality explanations it shows to the user (\textit{accepted set}), and maximize the ones for which it abstains (\textit{rejected set}). Thus, we report the percentage of low-quality explanations in the accepted and rejected sets when varying the rejection rate. We also report AUROC which assesses the rejector's ability to rank high-quality explanations above low-quality explanations. This metric is commonly used in novelty rejection~\cite{sun2022out}. 

\textbf{Setup}. We employ the following procedure: for each dataset, we (\textit{i}) train a linear SVM as the predictor $f$ on $\calT$ and compute the explanations and their corresponding machine-side metrics on a separate set $\calD$, (\textit{ii}) split $\calD$ into $\calD_{train}$, $\calD_{val}$ and $\calD_{test}$ ($40\%$/$10\%$/$50\%$), (\textit{iii}) fit the rejectors on $\calD_{train}$ and optimize their hyperparameters on $\calD_{val}$, (\textit{iv}) vary the rejection rate $\rho_\%$ from $1\%$ to $25\%$, and (\textit{v}) compute the metrics outlined in the previous paragraph on $\calD_{test}$. To improve robustness, we repeat steps (\textit{ii})--(\textit{v}) $10$ times and report the average results. All experiments were implemented in Python and executed on an Intel i7-12700 machine with 64 GB RAM. The experiments required approximately two days to complete.

\textbf{Implementation details}. Explanations are computed using \textit{KernelSHAP} \cite{lundberg2017unified} with $100$ samples and $\calT$ as background. We choose \textit{KernelSHAP} as it is one of the most well-known and widely used explainers~\cite{saarela2024recent}. To further support our findings, we also include results using \textit{LIME}~\cite{ribeiro2016should} in \cref{app:lime_experiment}. We include faithfulness and stability in \method's inputs since they are the most widely adopted machine-side metrics~\cite{quantus}, and train an \SVM as the rejector's classifier $\phi$ to assess explanation quality. Finally, we optimize \method's and the competitors' hyperparameters via grid search on $\calD_{val}$, see \cref{app:hyperparameters} for full details.

\subsection{Q1 and Q2: Benchmark Datasets}\label{sec:Q1-Q2}

\textbf{Datasets}. We evaluate all competitors simulating explanation quality judgments on \textit{five} datasets, comprising two medical ones and three standard benchmarks. The medical datasets cover distinct clinical tasks: {\tt iknl} is a synthetic breast cancer dataset where the goal is to predict the necessity of surgery for a given patient~\cite{iknl_synthetic_ncr}. {\tt sleepstage} focuses on distinguishing whether a subject is sleeping or awake based on EEG spectral characteristics~\cite{vallat2021open,hussain2022quantitative}. Finally, the remaining three datasets are widely used benchmarks~\cite{kelly2023uci} covering diverse application domains: {\tt adult}, {\tt compas}, and {\tt celeba}. Full details about the datasets and their preprocessing are provided in \cref{app:datasets}.

\textbf{Obtaining explanation quality judgments}. We obtain explanation quality judgments $y_{\vz} \in \{0,1\}$ by aggregating three distinct binary labels corresponding to faithfulness ($y^{faith}_{\vz}$), stability ($y^{stab}_{\vz}$), and plausibility ($y^{plaus}_{\vz}$). An explanation is deemed high-quality ($y_{\vz} = 1$) if it satisfies all three criteria: 
$$
    y_{\vz} = \left({y_{\vz}^{faith}} \; \wedge \; {y_{\vz}^{stab}} \; \wedge \; {y^{plaus}_{\vz}}\right).
$$
The low-quality labels for faithfulness ($y^{faith}=0$) and stability ($y^{stab}=0$) are assigned to the $u_{\%}$ least faithful and least stable explanations, while the remainder are considered high-quality. We set $u_{\%}$ to $10\%$ in our experiments. Other thresholds yield similar qualitative results (see \cref{app:different_thresholds}).

The plausibility labels $y^{plaus}$ are obtained by simulating human quality judgments. Specifically, we compare the explanations $\vz$ against ground-truth sets of task-relevant ($\calR$) and task-irrelevant ($\calI$) features. For the medical datasets, we build these following established clinical guidelines~\cite{shah2025can,berry2017aasm,hussain2022quantitative,vsuvsmakova2007classification}; while for the others, we define $\calI$ based on fairness constraints or possible spurious correlations. Then, an explanation $\vz$ is plausible ($y_{\vz}^{plaus} =1$) if its relevance scores correctly reflect the expected contribution for most of the features in $\calR$ and $\calI$. We simulate per-feature feedback similarly: we flag as incorrect the features whose scores contradict their actual relevance (\eg a feature in $\calI$ receives a high relevance score). Full details on how to simulate the plausibility labels can be found in \cref{app:featuresdatasets}. Moreover, we report an additional experiment in \cref{app:llm} where we obtain the simulated plausibility labels using an LLM.

\textbf{(Q1) Comparison with standard LtR strategies.}  
\begin{figure*}[!t]
  \centering
  \includegraphics[width=1\textwidth]{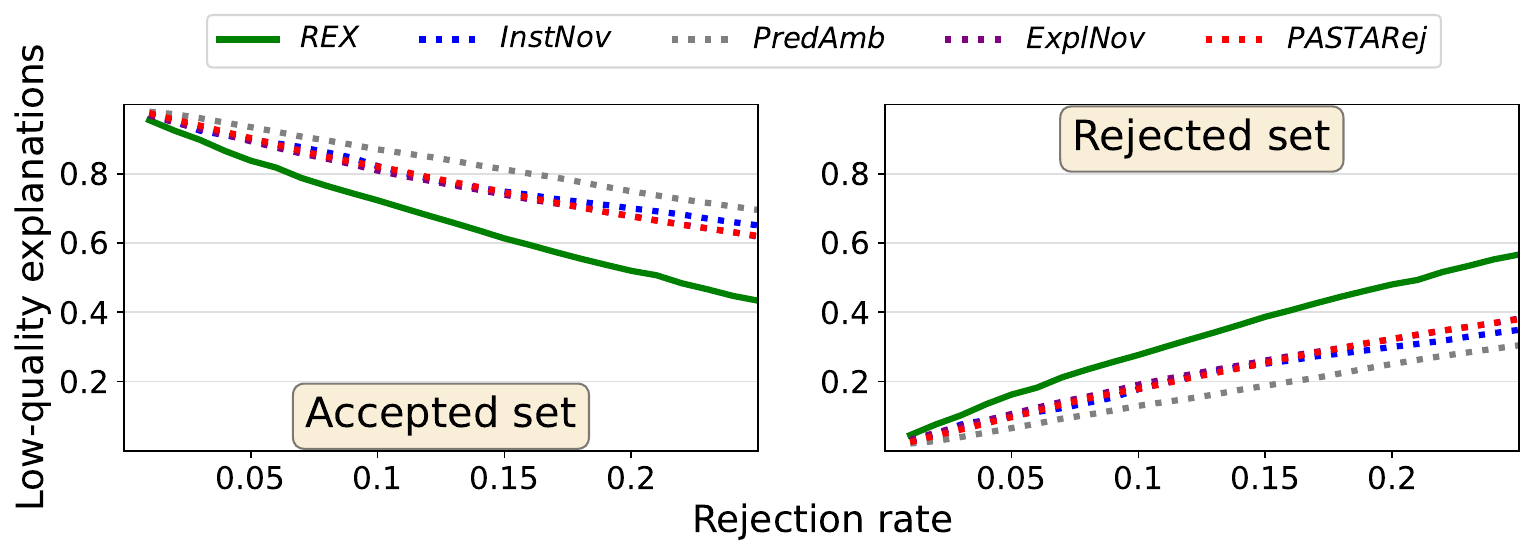}
  \caption{ 
  \textbf{\method rejects on average more low-quality explanations than all competitors.}
  Average percentage of low quality explanations in the accepted and rejected set for all the considered strategies over the five datasets for $25$ rejection rates $\rho_\%$. For all the considered rejection rates, \method consistently rejects more low-quality explanations than all competitors.}
  \label{fig:Q1}
\end{figure*}
\begin{table*}[!t]
    \caption{
  \textbf{\method outperforms its competitors at separating low-quality  from high-quality explanations on all datasets.}
  Average AUROC ($\pm$ std) for all the rejection strategies over the five datasets of the simulated setting. \method consistently obtains the best results in all cases.}
    \setlength{\tabcolsep}{4pt}
    \fontsize{9pt}{9pt}\selectfont
    \centering
    \begin{tabular}{l|ccccc}
    \toprule
    &  {\tt iknl} & {\tt sleepstage} & {\tt celeba} & {\tt adult} & {\tt compas} \\
    \midrule
    \method & \textbf{0.67 $\pm$ 0.02} & \textbf{0.77 $\pm$ 0.02} & \textbf{0.90 $\pm$ 0.01} & \textbf{0.94 $\pm$ 0.01} & \textbf{0.99 $\pm$ 0.01} \\
    \midrule
    \InstNov & 0.61 $\pm$ 0.01 & 0.54 $\pm$ 0.02 & 0.49 $\pm$ 0.02 & 0.63 $\pm$ 0.01 & 0.70 $\pm$ 0.02 \\
    \PredAmb & 0.62 $\pm$ 0.01 & 0.24 $\pm$ 0.01 & 0.69 $\pm$ 0.01 & 0.58 $\pm$ 0.01 & 0.56 $\pm$ 0.01 \\
    \ExplNov & 0.58 $\pm$ 0.01 & 0.57 $\pm$ 0.02 & 0.75 $\pm$ 0.01 & 0.68 $\pm$ 0.02 & 0.69 $\pm$ 0.01 \\
    \PASTARej & 0.60 $\pm$ 0.02 & 0.62 $\pm$ 0.07 & 0.65 $\pm$ 0.04 & 0.58 $\pm$ 0.06 & 0.75 $\pm$ 0.04 \\
    \bottomrule
\end{tabular}
    \label{tab:Q1}
\end{table*}
\cref{fig:Q1} shows the percentage of low-quality explanations for the accepted and rejected set as a function of the rejection rate averaged over the five considered datasets. On average, \method rejects more low-quality explanations than the competitors: about $11\%$ more than \InstNov, \ExplNov, and \PASTARej, and $17\%$ vs \PredAmb. Interestingly, even though \PASTARej aims at estimating the perceptual quality of an explanation, it perform similarly to other baselines, highlighting the need for a comprehensive score of explanation quality. Moreover, \method achieves the lowest number of accepted low-quality explanations in $78\%$ of the experiments (\ie across all (rejection rate, iteration) pairs), thus confirming the consistency of its performance.

\cref{tab:Q1} reports the average AUROC per dataset. In all datasets, \method is better at distinguishing between high- and low-quality explanations than its competitors, with 
an average improvement of at least $20\%$ against the best competitor \ExplNov. These results indicate that the classic LtR strategies such as instance novelty or prediction confidence are only weakly correlated with the quality of an explanation. This confirms the need for a rejection approach tailored to explanation quality. \ExplNov outperforms the standard LtR strategies. However, relying solely on the novelty of an explanation is insufficient to effectively identify low quality explanations. 

\textbf{(Q2) \method vs single metric-based rejectors.}
\begin{table*}[!t]
    \caption{
  \textbf{\method outperforms its ablated variants at separating low-quality from high-quality explanations.}
  Average AUROC ($\pm$ std) for \method and its ablated variants over the five datasets of the simulated setting. \method consistently obtains the best results in all datasets.}
    \setlength{\tabcolsep}{4pt}
    \fontsize{9pt}{9pt}\selectfont
    \centering
    \begin{tabular}{l|ccccc}
    \toprule
    & {\tt iknl} & {\tt sleepstage} & {\tt celeba} & {\tt adult} & {\tt compas} \\
    \midrule
    \method & \textbf{0.67 $\pm$ 0.02} & \textbf{0.77 $\pm$ 0.02} & \textbf{0.90 $\pm$ 0.01} & \textbf{0.94 $\pm$ 0.01} & \textbf{0.99 $\pm$ 0.01} \\
    \midrule
    \methodQ$_{faith}$ & \textbf{0.67 $\pm$ 0.02} & 0.67 $\pm$ 0.01 & 0.59 $\pm$ 0.01 & 0.61 $\pm$ 0.01 & 0.62 $\pm$ 0.01 \\
    \methodQ$_{stab}$ & 0.52 $\pm$ 0.01 & 0.57 $\pm$ 0.02 & 0.76 $\pm$ 0.01 & 0.69 $\pm$ 0.01 & 0.49 $\pm$ 0.01 \\
    \methodQ$_{human}$ & 0.58 $\pm$ 0.08 & 0.76 $\pm$ 0.03 & 0.76 $\pm$ 0.02 & 0.68 $\pm$ 0.04 & 0.86 $\pm$ 0.01 \\
    \bottomrule
\end{tabular}
    \label{tab:Q2}
\end{table*}
To investigate whether \method outperforms single-metric rejection strategies in assessing explanation quality, we evaluate three ablated variants of \method: \methodQ$_{faith}$, \methodQ$_{stab}$, and \methodQ$_{plaus}$, which reject explanations based solely on faithfulness, stability, and plausibility, respectively. For \methodQ$_{faith}$ and \methodQ$_{stab}$, we set the rejection threshold on the training set, and reject test explanations whose faithfulness or stability falls below the corresponding threshold. While, for \methodQ$_{plaus}$, we train a rejector identical to \method, but restrict its input solely to the explanations and its target to the plausibility labels. 

\cref{tab:Q2} reports the average AUROC per dataset for \method and its variants. Across all datasets, \method more effectively distinguishes between high- and low-quality explanations than its ablated variants, achieving an average improvement of at least $13\%$ against the best competitor \methodQ$_{plaus}$. 

\subsection{Q3: \method Predicts Human Judgments Better than LtR strategies}\label{sec:Q3}
Finally, we investigate whether \method can effectively mimic actual human judgments of explanation quality. While the previous research questions evaluated the combination of machine-side labels and simulated users, this experiment isolates the human dimension and aims to verify whether \method outperforms its competitors when using real user feedback. To this end, we collected high-quality human ratings of machine explanations through a large-scale annotation campaign, recruiting users with the crowd-sourcing platform Prolific  (\url{https://www.prolific.com}).\footnote{\textcolor{red}{The campaign was approved by our Research Ethics committee and Privacy office.}} Our study yielded annotated explanations for $1050$ predictions, with each explanation being annotated by five different labelers. 

\begin{figure}[!t]
    \centering
    \includegraphics[width=0.9\linewidth]{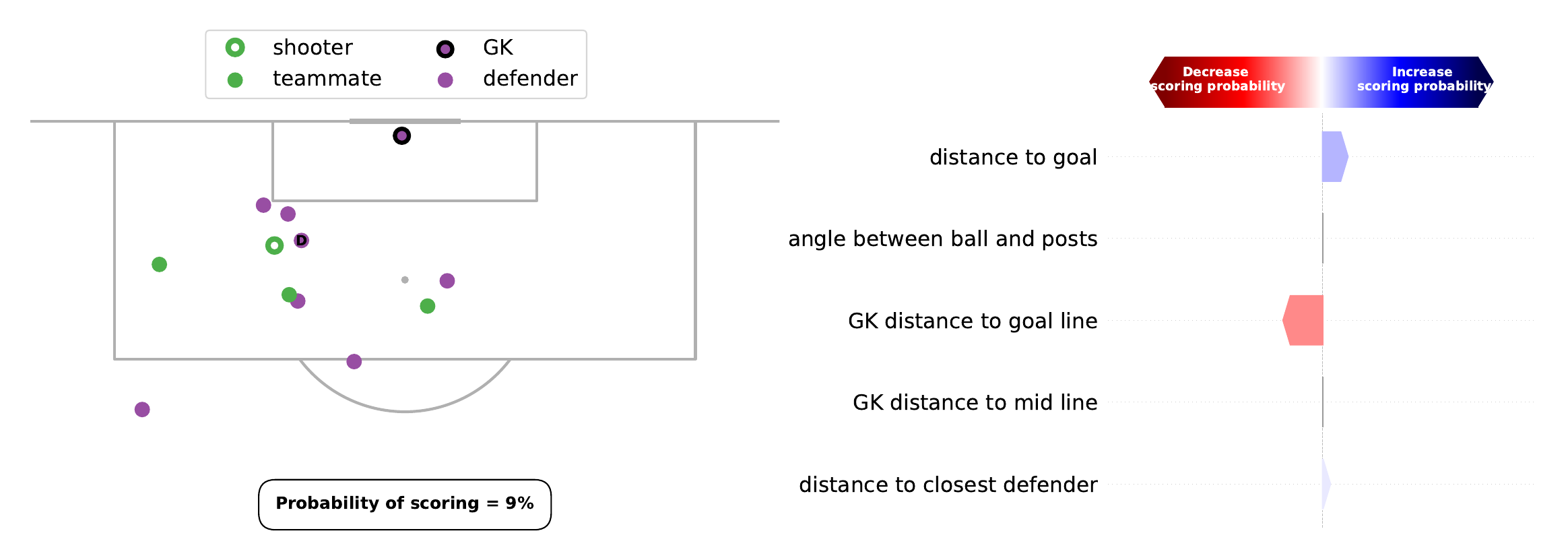}
    \caption{\textbf{Image from the user study} illustrating the snapshot (left), the predicted probability of scoring (bottom) and the associated Kernel\SHAP explanation (right). This suggests that the feature ``\textit{distance to goal}'' slightly increases the probability, while ``\textit{GK distance to goal line}'' decreases it.}
    \label{fig:snapshot}
\end{figure}

Our task was to explain the prediction of an expected goals (xG) model, which values the quality of a scoring opportunity in soccer as the probability that a shot results in a goal~\cite{xGmodel1}. Our choice stems from three considerations. First, Prolific enabled us to recruit subjects that possess the necessary domain expertise to perform the task, cf. \cref{app:human-annotation-process} for our vetting criteria. Second, all instances can be easily visualized, as shown in \cref{fig:snapshot}. Third, this is a real-world task with xG values being shown on TV and used in player recruitment~\cite{graham2024win}.\footnote{The model used in our experiments is not as complex as deployed models.} 

\textbf{Obtaining the explanations}. As a first step, we trained the predictor whose explanations we aim to annotate. Following standard practice in soccer analytics~\cite{xGmodel1}, we learned an XGBoost ensemble classifier~\cite{chen2016xgboost} to estimate the probability of a shot resulting in a goal. The training data consists of 21337 annotated shot events from the 2015-16 season in the top divisions of England, Spain, Germany and France~\cite{statsbomb2023data}.  For each shot, the location and the result (goal \textit{vs.} no goal) are recorded. Additionally, a snapshot is available, capturing the locations of the players visible in the broadcast video at the moment the shot is taken, cf. \cref{fig:snapshot} (left). From this data, we extract features that describe the positions of the shooter, goalkeeper, and nearest defender. Importantly, we include only features that are directly visualizable by the annotators in the snapshot. Explanations are generated on a separate set of 1050 shots from the 2015--16 season of the Italian top division on which the predictor achieves an AUROC of $0.81$. All preprocessing and training details are provided in \cref{app:obtaining-the-explanations}.

\textbf{Obtaining the annotations}.  Our goal is to obtain human-judgment labels on the explanation quality and per-feature feedback on the relevance scores. Given that subjective tasks are highly sensitive to interface design and question framing~\cite{stalans2012frames}, we designed our annotation protocol with the help of a psychologist and conducted several pilot studies to mitigate cognitive biases~\cite{bertrand2022cognitive}. Participants ($N=175$) were recruited via Prolific while annotations were collected through Google Forms. Each participant annotated $30$ trials. In each trial, participants were shown a snapshot of a shot and the corresponding prediction and explanation, cf. \cref{fig:snapshot}. The left side shows the position of all involved players and the ball, along with the model's prediction. The right side shows the relevance scores of each feature as arrows indicating whether the feature increases or decreases the predicted probability of scoring. The features were chosen specifically to be easily interpretable and visually grounded, enabling intuitive assessment by the annotators. These were requested to specify how much they agreed with the model's prediction and, separately, with its explanation using two $5$-point Likert-scale questions ($1 =$\; completely disagree, $5 =$\;completely agree).  Next, they were asked to optionally select individual features they believed were misused in the explanation, \ie had an incorrect relevance score, via a multiple-choice question. We validated our experimental design by tracking the consistency of individual annotations in two pilot studies: on average, annotators tended to assign consistent scores to the same explanation across repeated trials. Full details about our procedure are provided in \cref{app:human-annotation-process}.

\textbf{Annotation preprocessing}. To ensure high-quality annotations, we filtered out participants that failed an attention check, rated all explanations the same, or did not flag any as incorrect, leaving us with $149$ participants. Additionally, because evaluating explanations is highly subjective, we removed explanations with low inter-annotator agreement. We aggregated the explanation scores using the average and considered explanations with an average score lower than 3 as low-quality, and the others as high-quality~\cite{joshi2015likert,batterton2017likert}. For feature-level feedback, we marked a relevance score as incorrect if the majority of annotators agreed that the corresponding feature was misused.

\textbf{Results.} 
\begin{figure*}[!t]
  \centering
  \includegraphics[width=1\textwidth]{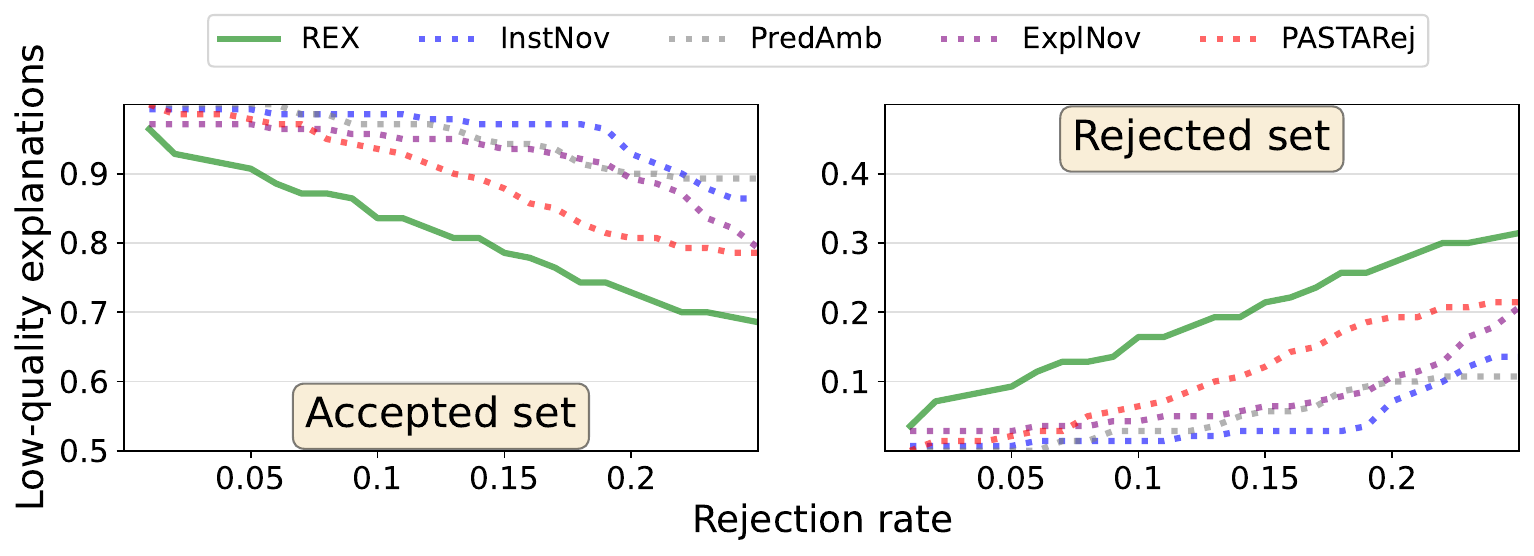}
  \caption{ 
  \textbf{\method rejects on average more low-quality explanations than all competitors over the user study data.}
  Average percentage of low quality explanations in the accepted and rejected set for all the considered strategies over the user study data for $25$ rejection rates $\rho_\%$. For all the considered rejection rates, \method consistently rejects more low-quality explanations than all competitors.}
  \label{fig:Q3}
\end{figure*}
\begin{table*}[!t]
    \caption{
  \textbf{\method outperforms its competitors also when the explanation  quality labels are obtained through real users.}
  Average AUROC ($\pm$ std) for \method, standard LtR strategies, and \PASTARej over the user study data.}
    \setlength{\tabcolsep}{4pt}
    \fontsize{9pt}{9pt}\selectfont
    \centering
    \begin{tabular}{l|c|cccc}
    \toprule
    & \method & \PredAmb & \InstNov & \ExplNov & \PASTARej \\
    \midrule
    user study & \textbf{0.64 $\pm$ 0.05} & 0.46 $\pm$ 0.07 & 0.39 $\pm$ 0.07  & 0.49 $\pm$ 0.06 & 0.53 $\pm$ 0.05 \\
    \bottomrule
\end{tabular}
    \label{tab:Q3}
\end{table*}
\cref{fig:Q3} shows the percentage of low-quality explanations for the accepted and rejected set as a function of the rejection rate for the user study data. \method consistently outperforms the competitors by achieving a lower proportion of accepted low-quality explanations across all rejection rates.

\cref{tab:Q3} reports the average AUROC for \method and its competitors, measuring their ability to distinguish between high-quality and low-quality explanations. \method outperforms all competitors, achieving at least an $11\%$ improvement in AUROC. Notably, we observe that the classic LtR strategies perform similarly to random, thus confirming that these are unsuited to assess explanation quality also when humans provide the quality judgments. To evaluate statistical  significance, we apply the method proposed in~\cite{demvsar2006statistical}. A one-way ANOVA test rejects the null-hypothesis that all methods perform similarly. According to the Tukey’s HSD~\cite{demvsar2006statistical}, \method offers significantly better performance than the baselines; the corresponding $95\%$ confidence interval plot is provided in \cref{fig:Q1ci} in \cref{app:Q1ci}. 

Finally, we found that human annotators identified, on average, 1.8 features with incorrect relevance scores in low-quality explanations, compared to only 0.7 features in high-quality ones. This supports our intuition that users perceive low-quality explanations as containing more wrong relevance scores.

%% file: texFiles/5_RelatedWork.tex
\section{Related Work}\label{sec:related-work}
\textbf{Learning to Reject}.  The problem of deferring hard decisions has been studied in the context of \textit{learning to reject}, \textit{learning to defer} \cite{mozannar2020consistent}, \textit{learning under algorithmic triage} \cite{okati2021differentiable}, \textit{learning under human assistance} \cite{de2021classification}, and \textit{learning to complement} \cite{bansal2021most}; see \cite{hendrickx2024machine} for a recent survey. These approaches all enable the machine to offload certain decisions to a human expert, but differ in what criterion they use. While some strategies entirely rely on the machine's self-assessed uncertainty~\cite{cortes2016learning,pugnana2023model}, others implement the rejection policy as a machine learning classifier and optimize it for joint team performance~\cite{madras2018predict} or learn the classifier and the policy jointly~\cite{wilder2021learning}. None of them, however, considers the role of explanations in decision making, which we argue is central.

\textbf{Explainable AI} (XAI) aims at designing mechanisms for properly justifying algorithmic decisions to end-users in non-technical terms~\cite{adadi2018peeking}. We focus on (post-hoc) feature attribution techniques, which highlight what features influenced a prediction the most. Many high profile techniques belong to this group, \eg\ \LIME \cite{ribeiro2016should}, \SHAP \cite{datta2016algorithmic,lundberg2017unified}, input gradients \cite{sundararajan2017axiomatic}, and formal feature attributions \cite{yu2023formal}. With respect to feature-attribution methods, \method is explainer-agnostic, \ie it can assess the quality of explanations irrespectively of how these are computed. The only work that combines XAI and LtR is \cite{artelt2023not}, which focuses on explaining the reasons behind rejection using counterfactuals, and as such is orthogonal to our work.

%% file: texFiles/6_Conclusion.tex
\section{Conclusion}\label{sec:conclusion}
We have introduced the problem of \textit{learning to reject low-quality explanations} (LtX) and proposed \method, a simple yet effective technique for learning a rejector from a limited amount of explanation-quality labels. Our empirical analysis showcases how, in contrast to standard LtR approaches,  \method successfully identifies low-quality explanations in both synthetic and human-annotated tasks. In future work, we will extend our setup to learn the rejector and classifier jointly, so as to optimize their overall performance \cite{de2021classification,wilder2021learning}, and look into leveraging \method's rejector for debiasing confounded ML models by rating their explanations \cite{teso2023leveraging}.

\textbf{Limitations.} \method is designed to identify and offload predictions associated with unsatisfactory explanations and does not optimize predictive performance. However, \method can be combined with state-of-the-art LtR strategies to ensure \textit{both} incorrect predictions and unsatisfactory explanations are rejected.  Another limitation of \method is that, just like \PASTA~\cite{kazmierczak2024benchmarking}, it relies on high-quality human annotations. We argue that this is necessary in high-stakes applications, but also that good annotations are likely to be available anyhow as in these settings expert users \textit{have} to oversee machine decisions at all times~\cite{hoffman2018metrics,zhou2021evaluating}.

%% file: texFiles/Appendix.tex
\newpage

\section{Explanation quality metrics}\label{app:explanation_metrics}
Explanation quality metrics aim to assess to what extent explanations satisfy the general goal of explaining a decision. These metrics can be broadly categorized into two  families~\cite{lopes2022xai,zhou2021evaluating}: \textit{machine-side} and \textit{human-side} metrics. The former focus exclusively on the relationship between the explainer and the predictor, whereas the latter involve human subjects in evaluating the quality of the explanations.

\subsection{Machine-side metrics.}\label{app:machineside}
The simplest way to evaluate an explanation is by verifying whether it effectively reveals the predictor's underlying reasoning. Several metrics have been proposed to assess the relationship between explanations and the predictor. \cite{chen2022makes} categorize existing machine-side metrics - and provide their mathematical formulations — into three groups: stability, faithfulness, and complexity. We exclude homogeneity from our analysis because it is defined for groups of explanations rather than individual ones.

\textbf{Stability} measures the similarity of explanations under changes to the input example, the training data, or the model hyperparameters~\cite{kalousis2007stability}. This can be harmful because an attacker can selectively choose explanations based on their (potentially adversarial) interests~\cite{schneider2023deceptive}. Following~\cite{bansal2020sam}, we define the stability of an explanation as the average similarity across multiple runs of the same explainer, each potentially yielding a different explanation. Formally, given an example $\vx$ and prediction $f(\vx)$ with associated explanation $\vz$, \textit{stability} is defined as:
\begin{equation}
        \operatorname{stab}(\vz) = \mathbb{E}_{\vz^{\prime} \sim \calZ}\left[Sim(\vz, \vz^{\prime})\right]
    \label{eq:uniqueness}
\end{equation}
where $Sim$ is a similarity metric and $\calZ$ denotes the space of possible explanations for the given prediction. In practice, we compute stability using the Pearson correlation coefficient as the similarity metric and average it across ten independently generated explanations. 

\textbf{Faithfulness} measures how accurately an explanation captures the true underlying behavior of the predictor~\cite{dasgupta2022framework}. Given an explanation $\vz$, we define the sets of relevant features $\vz_{\calR} = \{ i < d : |\vz_{i}| > 0\}$ and irrelevant features $\vz_{\calI} = \{ i < d : |\vz_{i}| = 0\}$. Intuitively, an explanation is faithful if perturbing irrelevant features causes little to no change in the predictor's output, while perturbing relevant features induces significant changes. Building on \cite{azzolin2025reconsidering}, we define \textit{faithfulness} ($\operatorname{faith}$) as the harmonic mean of \textit{sufficiency} ($\operatorname{suf}$) and \textit{necessity} ($\operatorname{nec}$), which estimate the sensitivity of the prediction to perturbations in irrelevant and relevant features, respectively. Formally, given a example-prediction pair $\left(\vx, f(\vx)\right)$ with associated explanation $\vz$, and the predictor to be explained $f$, \textit{sufficiency} and \textit{necessity} are defined as:
\begin{equation}
\operatorname{suf}_{d, p_\calI}\left(\vz\right)=\mathbb{E}_{\vx^{\prime} \sim p_\calI}\left[\Delta_f\left(\vx, \vx^{\prime}\right)\right]
\label{eq:sufficiency}
\end{equation}
\begin{equation}
    \operatorname{nec}_{d, p_\calR}\left(\vz\right)=\mathbb{E}_{\vx^{\prime} \sim p_\calR}\left[\Delta_f\left(\vx, \vx^{\prime}\right)\right]
\label{eq:necessity}
\end{equation}
where $\Delta_f$ measures prediction change between $\vx$ and its perturbed version $\vx^{\prime}$, and $p_{\calR}$ and $p_{\calI}$ are interventional distributions that specify how to perturb relevant and irrelevant features, respectively. \cref{eq:sufficiency} and \cref{eq:necessity} are then normalized to $[0,1]$ range, the higher the better, via a non-linear transformation \ie respectively $exp\left(-\operatorname{suf}_{d, p_\calI}\right)$ and $1 - exp\left(-\operatorname{nec}_{d, p_\calR}\right)$. Operationally, for a given example-explanation pair $(\vx, \vz)$ sampling from $p_{\calR}$ ($p_{\calI}$) involves perturbing the features in $\vz_{\calR}$ ($\vz_{\calI}$) following \cite{caruana2006compression}, while keeping the remaining features fixed. Additionally, the prediction change $\Delta_f$ is computed either as the absolute difference in positive class probability for classification tasks, \ie $|P(Y=1 | \vx) - P(Y=1 | \vx^{\prime})|$, or the absolute prediction difference in regression, \ie $|f(\vx) - f(\vx^{\prime})|$.

\textbf{Complexity} refers to the cognitive burden associated with parsing an explanation~\cite{chalasani2020concise}. In general, a less complex explanation is easier for a human to understand, making complexity a common proxy for understandability~\cite{molnar2020interpretable}. Following \cite{bhatt2020evaluating}, given an example $\vx$ with prediction $f(\vx)$ and explanation $\vz$, the \textit{complexity}  can be computed as:
\begin{equation}
\operatorname{compl}_{d, p_\calI} = \mathbb{E}\left[-\ln \left(\overline{\vz}\right)\right]=-\sum_{i=1}^d \overline{\vz_{i}} \ln \left(\overline{\vz_{i}}\right)
\end{equation}
where $\overline{\vz_{i}}$ is the fractional contribution of feature $i$, \ie the ratio of its absolute relevance score $\left|\vz_{i}\right|$ to the sum of all the absolute relevance scores $\sum_{j=1}^d \mid \vz_{j} \mid$. We exclude complexity from our experimental evaluation, as we inherently limit the number of non-zero relevance scores, making its impact on quality marginal in our experiments.

\subsection{Human-side metrics}\label{app:humanside}
Despite the literature recognizing the importance of human-centered evaluations~\cite{kazmierczak2024benchmarking}, only a few metrics have been proposed to evaluate explanations from the perspective of a human~\cite{naveed2024overview}. This gap stems from the inherently subjective nature of human evaluations, which typically makes it challenging to provide a precise mathematical formulation for a metric~\cite{chen2022makes}. Moreover, there is no consensus in the literature regarding standard criteria for human-side evaluation metrics~\cite{zhou2021evaluating}.

\textbf{PASTA} uses a model to score each explanation based on how this is perceived by humans~\cite{kazmierczak2024benchmarking}. The authors first construct a dataset in which users rated several explanations according to four key desiderata: faithfulness, robustness, complexity, and objectivity. Then, the \textit{PASTA-metric} is trained on these ratings to derive a metric value for new explanations. Specifically, this model consists of two main components: an embedding network that leverages a foundation model to generate feature embeddings from the explanations, and a scoring network that employs a linear layer to predict the human ratings based on these embeddings. \PASTA is the closest competitor to our work, as it also aims to assess explanations' quality. However, there are three substantial differences with our approach. First, \PASTA solely evaluates explanations' quality from the human perspective, whereas we aim to score the overall quality of an explanation by considering both the machine- and the human-side. Second, \PASTA is designed for image data and relies on an embedding network to create embeddings from this high-dimensional space; in contrast we focus on tabular data and learn directly from feature-importance explanations. Third, \PASTA seeks to create a dataset-agnostic metric and thus annotates $25$ explanations per dataset to encourage generalization. Conversely, we aim to train a dataset-specific rejector and therefore collect $1050$ annotations for a single dataset.

\textbf{Other human-side metrics.} \textit{\textbf{Understandability}} measures whether an explanation is easy to comprehend for the human~\cite{lopes2022xai}. The rationale behind this metric is to examine whether the explanations facilitate the user's understanding of the model's decisions~\cite{dieber2022novel}.
\textit{\textbf{Plausibility}} is high if $\vz$ matches the ground-truth explanation $\vz^*$, assuming the latter exists and is unique~\cite{naveed2024overview,jin2024evaluating}. Depending on the model's behavior and structure of the underlying learning problem, the model's reasoning may or may not reflect the ground-truth explanation $\vz^*$. 

\section{User study}\label{app:userstudy}

\subsection{Data} For this user study, we used the publicly available StatsBomb 360 event stream data~\cite{statsbomb2023data}. This contextualized event stream data is extracted from broadcast video and contains event stream data, and snapshots of player positioning at the moment of each event. The event stream data describes semantic information about the on-the-ball actions, such as which actions are performed, their start and end location, the outcome of the action, which players performed them, and the time in the match they were performed at.

\subsection{Obtaining the explanations}\label{app:obtaining-the-explanations}
To obtain the explanations, we begin by preprocessing the data to obtain the features needed to train the classifier. From each shot snapshot, we extract the following features: (i) the distance from the ball to the center of the goal, (ii) the angle between the ball and the goalposts, (iii) the distance of the goalkeeper from the goal line, (iv) the distance of the goalkeeper from the midline (\ie the line that passes through the center of the field and the middle of the goals), and (v) the distance to the closest defender (excluding the goalkeeper). We select only these features for two main reasons: they are easily interpretable from the snapshot (see \cref{fig:snapshot}), and their meanings are non-overlapping, which makes it easier for annotators to disentangle their individual contributions as we found empirically that working with strongly correlated features can complicate human assessment. Using these features, we train an XGBoost ensemble~\cite{chen2016xgboost} consisting of $50$ trees with a maximum depth of $3$, as it is standard practice in soccer analytics~\cite{xGmodel1}. The model is trained on shots from the 2015–2016 season across four major top-tier leagues (Germany, Spain, England, and France). We evaluate the classifier on a held-out test set of 1,050 shots from the Italian top division in the same season. The primary goal of xG is to produce well-calibrated probability estimates because they are used for decision making (\eg evaluating players and giving advice about when to shot), which we assess by reporting the Brier score. Additionally, goals should receive an higher scoring probability than non-goals, which we capture by using AUROC. The model achieves a Brier score of $0.067$ and an AUROC of $0.81$.

We then use the test set to generate the explanations. As for the benchmark datasets, explanations are generated using Kernel\SHAP\cite{lundberg2017unified} with 100 samples and the training set used as background.

\subsection{Human annotation process}\label{app:human-annotation-process}
Participants were recruited using Prolific, a crowd sourcing platform. We applied Prolific's filters to ensure that participants possessed sufficient soccer expertise. Specifically, we applied filters to recruit subjects that (\textit{i}) live in countries where soccer is widespread (UK, Germany, France, Spain, Belgium, Italy, Netherlands, or Portugal), and (\textit{ii}) actively watch and play soccer. All participants were compensated with £3 for an expected completion time of 25 minutes, as estimated from the pilot studies.

After conducting pilot studies to ensure that the task was clear and comprehensible and to verify intra-annotator consistency, we launched the main user study. Participants were first requested to give their consent to participate. Then, they were provided with a link to an external Google Doc containing task instructions, which they could consult at any time during the session. The document provides general introduction for the task setting and objective, the description and illustration of the predictor's features, and 3 exemplary snapshots. 

\begin{figure}
    \centering
    \includegraphics[width=0.5\linewidth]{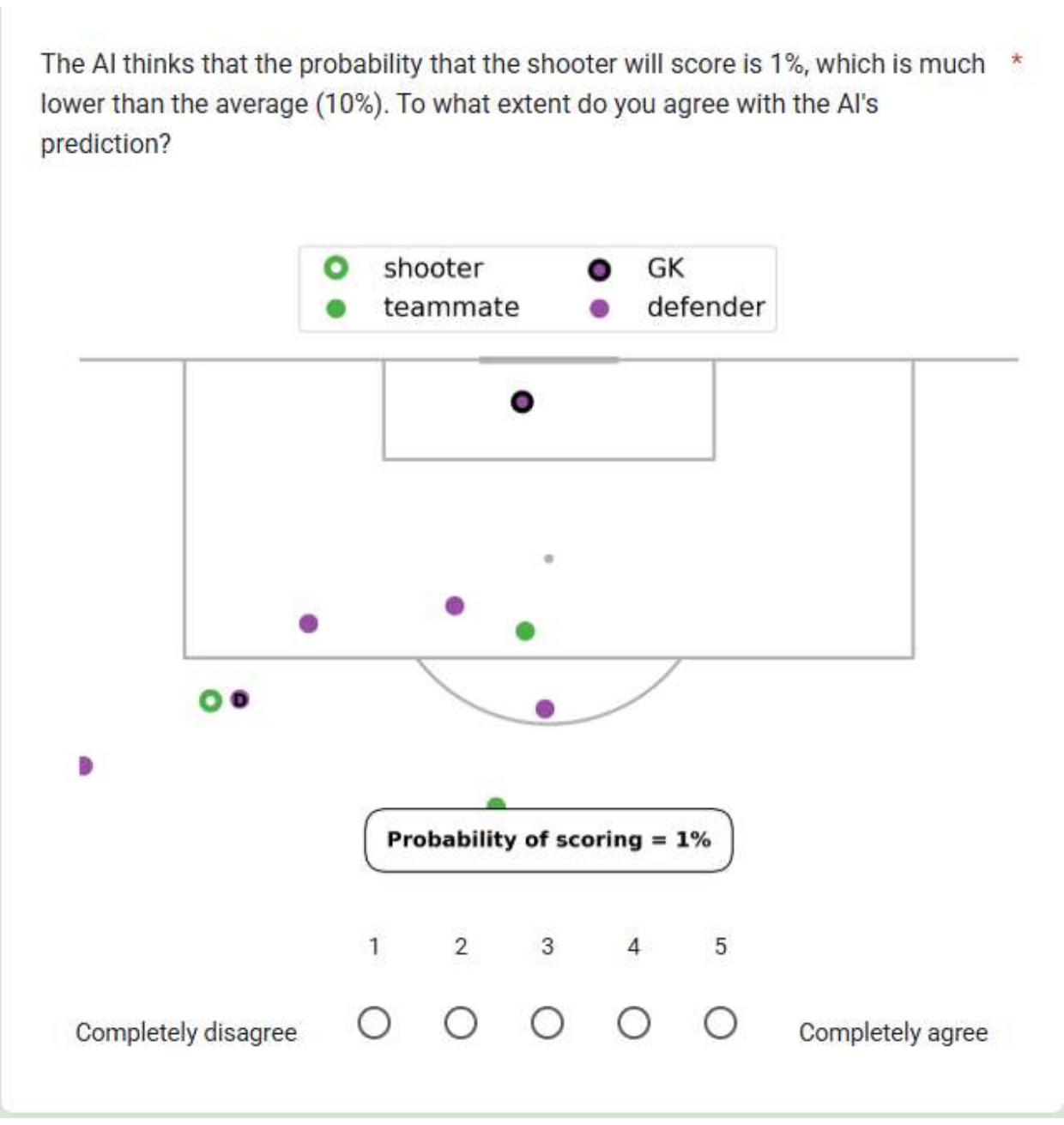}
    \caption{Example of the first image of each trial}
    \label{fig:prediction-snapshot}
\end{figure}

\begin{figure}
    \centering
    \includegraphics[width=0.5\linewidth]{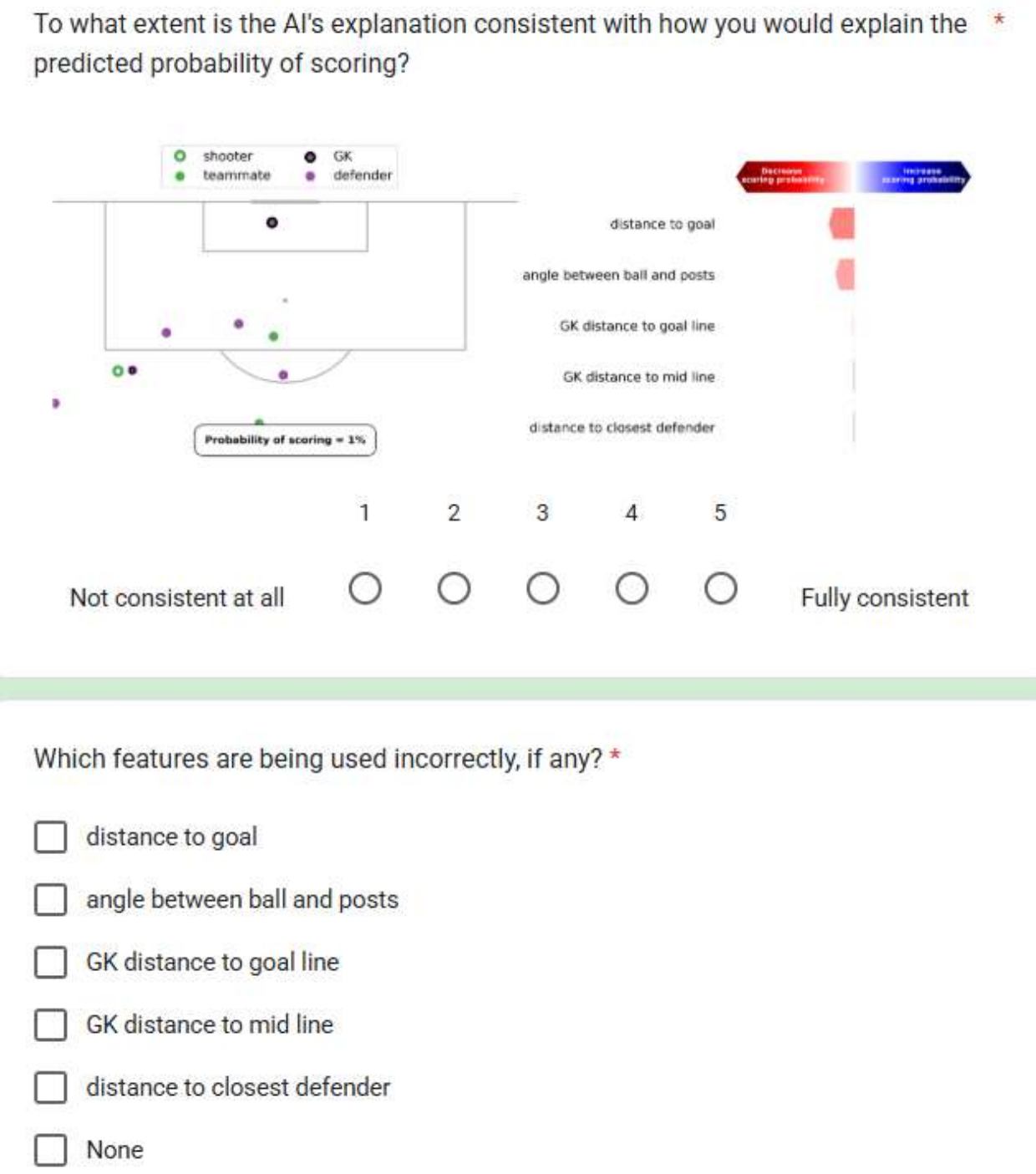}
    \caption{Example of the second image of each trial}
    \label{fig:explanation-snapshot}
\end{figure}

After the task introduction, participants completed three warm-up trials to familiarize themselves with the interface and the task; this was followed by the real annotation session comprising of 30 trials. In each trial, participants were asked three questions: two 5-point Likert-scale questions to separately assess the quality of the prediction and explanation, and one multiple choice question to identify the features with a wrong relevance score. We used two separate questions, presented in distinct sections of the form, to disentangle participants' agreement with the prediction from their perception of the explanation's quality and to minimize spurious correlations between their responses. 5-point Likert scales have been chosen as they provide satisfactory reliability and validity~\cite{taherdoost2019best}. Specifically, in the first question, participants were shown an image containing only the shot snapshot along with the predicted probability of scoring (see \cref{fig:prediction-snapshot}) and asked to assess their agreement with the prediction - \textit{"The AI thinks that the probability that the shooter will score is 1\%, which is much lower than the average (10\%). To what extent do you agree with the AI's prediction?"}, where the comparison \textit{much lower} was dynamically adapted based on the predicted probability. For the second and third questions, participants were shown a different image containing the shot snapshot, the prediction, and the explanation (see \cref{fig:explanation-snapshot}). To facilitate interpretation, features relevance are visualized as independent arrows: blue indicates a positive impact on the prediction, while red indicates a negative impact. The second question - \textit{"To what extent is the AI's explanation consistent with how you would explain the predicted probability of scoring?"} - was used to collect the perceived explanation quality. While the third question - \textit{"Which features are being used incorrectly, if any?"} - is used to obtain the feature-level feedback about the features with an incorrect relevance score in the explanation. To ensure high-quality annotations, we included an attention check requiring specific answers for a trial. This allowed us to detect and discard inattentive or randomly answering participants.

\subsection{Annotations preprocessing}
To ensure high-quality annotations, we applied several filtering steps. First, we excluded participants who failed more than one attention check question, as well as those who consistently provided the same score for every explanation (typically a score of 3), since this means they were not able (or did not bother) to discriminate between explanations. We also removed two participants who did not flag any relevance score as incorrect. Additionally, given the subjective nature of the task (for instance, we saw that some explanations showed very low annotator agreement, \eg 1 vs 5) we removed explanations for which the standard deviation of the explanation quality scores exceeded $1.25$. This step helped ensuring that our dataset contains only explanations where annotators' opinions are reasonably consistent. After applying these filters, $718$ explanations remained for our experiments.

\section{Experiments: extended details and results}
\label{app:experiments}

\subsection{Dataset characteristics and predictor's performance}\label{app:datasets}
\cref{tab:datasets} presents the characteristics of the five datasets used in the empirical evaluation, along with the performance of the predictor $f$. Specifically, a linear SVM is trained as predictor $f$  on a training set $\calT$ and evaluated on a test set $\calD$. The size of $\calD$ is limited because obtaining human-judgment labels on explanation quality is expensive~\cite{kazmierczak2024benchmarking}. Since some datasets are unbalanced, we report the balanced accuracy (\textit{BACC}) as performance metric. Additionally, the table reports the proportion of low-quality explanations $\gamma$ in $\calD$ for each dataset, as determined using the procedure described in \cref{sec:Q1-Q2}. Detailed descriptions of medical dataset and specific preprocessing steps are provided in the following paragraphs.
\begin{table*}[!t]   
\setlength{\tabcolsep}{5pt}
    \caption{\textbf{Datasets' characteristics and predictor's performance.}
    This table reports the datasets' characteristics (\ie size of the training set $\#$($\calT$), number of features $d$, size of the test set $\#$($\calD$), proportion of low-quality explanations $\gamma$) and the predictor $f$'s performance on the five datasets used in the simulated experiments.}
    \centering
    \begin{tabular}{l|cccc|c}
    \toprule
     dataset & $\#$($\calT$) & $d$ & $\#$($\calD$) & \textit{BACC}$_f$ $\uparrow$  & $\gamma$\\
     \midrule
     {\tt iknl} & 10000 & 16 & 2000 & 0.553 & 0.33 \\
     {\tt sleepstage} & 100 & 21 & 962 & 0.708 & 0.60\\
     {\tt celeba} & 10000 & 39 & 2000 & 0.929 & 0.33\\
     {\tt adult} & 10000 & 12 & 2000 & 0.757 & 0.27\\
     {\tt compas} & 10000 & 12 & 2000 & 0.690 & 0.43\\
    \bottomrule
    \end{tabular}
    \label{tab:datasets}
\end{table*}

\textbf{{\tt iknl}}. The synthetic dataset {\tt iknl} comprises breast cancer data for 60,000 patients spanning the years 2010 to 2019~\cite{iknl_synthetic_ncr}. All patients with missing values are removed, leaving a total of 38846 patients. The dataset contains 46 features, including five potential target variables: chemotherapy, hormonal therapy, radiotherapy, targeted therapy, and surgery.  In this paper, we focus exclusively on predicting whether a given patient will require \textit{surgery} or not. The data is highly imbalanced; \ie only $10\%$ of patients undergo surgery.

\textbf{{\tt sleepstage}}. We retrieved and preprocessed the data following the protocols of the YASA library~\cite{vallat2021open}. From each PSG night, we extracted a single central EEG channel. These signals were downsampled to $100$~Hz to reduce computational cost and bandpass-filtered between $0.40$~Hz and $30$~Hz. We then computed frequency-domain features from the signal's periodogram using 30-second epochs. These features included the absolute power of the broadband signal, specific power ratios (delta/theta, delta/sigma, delta/beta, alpha/theta), and the relative spectral power in the following bands: slow ($0.4$-$1$~Hz), delta ($1$–$4$~Hz), theta ($4$–$8$~Hz), alpha ($8$–$12$~Hz), sigma ($12$–$16$~Hz), and beta ($16$–$30$~Hz). To incorporate temporal context, we applied a smoothing strategy: features were smoothed using a rolling average over the preceding 2 minutes. Finally, we simplified the classification task into a binary problem, aggregating stages N1, N2, N3, and REM into the "Sleep" class, while labeling the remaining stage as "Wake".

\subsection{Simulating the plausibility labels}\label{app:featuresdatasets} 
To obtain the simulated plausibility labels, we start defining the sets of task-relevant ($\calR$) and task-irrelevant features ($\calI$) for each dataset. For {\tt iknl}, we follow the Oncoguide-2020 guidelines~\cite{iknl_synthetic_ncr}, defining the set of relevant features $\calR$ as \{\textit{cstadium}, \textit{morf}, \textit{ct}, \textit{cm}, \textit{cn}, \textit{er\_stat}, \textit{pr\_stat}\}, while \textit{dcis\_comp} is deemed as an irrelevant feature in \calI~\cite{shah2025can}. For {\tt sleepstage}, the set $\calR$ includes the power ratios alpha/theta, delta/theta, delta/sigma, and delta/beta. These were selected as they are established biomarkers for discriminating between sleep and wake stages~\cite{vsuvsmakova2007classification,hussain2022quantitative}. For the remaining datasets, we define $\calI$ based on fairness constraints or possible spurious correlations. Specifically, in {\tt celeba}, we include known spurious features that exploit known dataset biases, \ie \textit{smiling}, \textit{eyeglasses}, and \textit{blond-hair}, as these act as spurious proxies rather than true indicators of gender~\cite{gat2022latent,hand2018doing}. To prevent discriminatory reasoning in {\tt adult} and {\tt compas}, we assign protected attributes to $\calI$: \textit{sex}, \textit{race}, and \textit{native-country} for {\tt adult}, and \textit{sex} and \textit{race} for {\tt compas}. \\ 
Then, for each explanation $\vz$, we obtain the simulated plausibility label $y^{plaus}_{\vz}$ by computing its overlap with $\calR$ and $\calI$. Let $g \in \left\{-1, 0, 1\right\}^d$ be a ground-truth vector where $g_i = 1$ if feature $i \in \calR$, $g_i=-1$ if $i \in \calI$, and $g_i=0$ otherwise. Following~\cite{shah2025can}, we define the alignment indicator $\delta(\cdot)$ for the $i$-th feature as:
\begin{equation}
\delta(\vz_i,\vg_i)=
\begin{cases}
1 & \text { if }  \left(\vz_i \ne 0 \text { and } \vg_i=1\right) \text { or } \left(\vz_i=0 \text { and } \vg_i=-1\right) \\  
0 & \qquad \qquad \qquad \text { otherwise }
\end{cases}
\end{equation}
The alignment score $\alpha\left(\vz,\vg\right)$ is then obtained as:
\begin{equation}
\alpha(\vz, \vg)=\frac{1}{\|\vg\|_0} \sum_{i=1}^{|\vg|} \delta(\vz_i, \vg_i)
\end{equation}
Finally, an explanation is plausible from the human perspective ($y^{plaus}_{\vz} =1$) if $\alpha\left(\vz,\vg\right) > \kappa$, and low-quality otherwise. In our experiments, we set $\kappa = 0.5$. Finally, to simulate the per-feature labels, we identify the features with incorrect relevance scores using the same alignment indicator. A relevance score for a feature $i$ is considered wrong ($i \in \calW_{\vz}$) if it is contained in $\calR$ or $\calI$ but fails to align with them. Formally,
$
\calW_{\vz} = \left\{i \in \{1, \dots, d\} \mid \vg_i \ne 0 \wedge \delta(\vz_i, \vg_i) = 0\right\}
$
and 
$
\calC_{\vz} = \left\{i \in \{1, \dots, d\} \mid \vg_i \ne 0 \wedge \delta(\vz_i, \vg_i) = 1\right\}
$.

\subsection{Hyperparameter Selection}\label{app:hyperparameters}
We optimize all hyperparameters using a grid search on the validation split $\calD_{val}$. Specifically, for \method we optimize the \SVM kernel (linear, polynomial, RBF), the cost of mistakes $C \in \{0.1, 1, 10\}$,  the number of augmentations per explanation $k \in \{5,10,20\}$ and the noise $\epsilon_0 \in \{0.1, 0.5, 1\}$. For \PASTARej, we employ the authors' code for the scoring network and optimize the loss hyperparameters $\alpha \in \{0.1, 1, 10\}$, $\beta \in \{0.001, 0.01, 0.1\}$ and $\gamma \in \{0.01, 0.1, 1\}$. For \InstNov and \ExplNov, we optimize the number of neighbors $k_{NN} \in \{1, 5, 10\}$.

\subsection{\method's input space}\label{app:input_space} 
To investigate which inputs the rejector needs to assess explanation quality, we consider three variants of \method that uses additional input for the rejector along with the explanation and the machine-side metrics: \methodQ$_{X}$ also uses its corresponding example, \methodQ$_{Y}$ also uses prediction, and \methodQ$_{X,Y}$ uses both the example and the prediction. For each variant, we augment the explanations  (see~\cref{sec:creating_dataset}), and train the rejector on a training set obtained by concatenating each (augmented) explanation with the metric and the example, the prediction, or both.

\cref{tab:Q2app} reports the average AUROC per dataset for \method and each of the above variants. Interestingly, including the examples as part of the rejector’s input tends to decrease the performance due to the limited number of explanation quality labels which makes it difficult for the rejector to learn the relationship between the explanations and the examples. Moreover, even concatenating only the prediction as in \methodQ$_{Y}$ results in a small performance hit (on average $2\%$), suggesting that explanations alone are often sufficient.

\begin{table*}[!t]
    \caption{
    \textbf{\method outperforms its variants that additionally provide examples and/or predictions as input to the rejector.}
    Average AUROC for \method and three variants using different inputs to learn the quality of an explanation over the five datasets. 
    \method consistently achieves the highest AUROC across all datasets, showing that explanations alone suffice for the rejector to assess their quality.
    }
\setlength{\tabcolsep}{4pt}
\fontsize{8pt}{8pt}\selectfont
    \centering
\begin{tabular}{l|ccccc}
    \toprule
    & {\tt iknl} & {\tt sleepstage} & {\tt celeba} & {\tt adult} & {\tt compas} \\
    \midrule
    \method & \textbf{0.67 $\pm$ 0.02} & \textbf{0.77 $\pm$ 0.02} & \textbf{0.90 $\pm$ 0.01} & \textbf{0.94 $\pm$ 0.01} & \textbf{0.99 $\pm$ 0.01} \\
    \midrule
    \methodQ$_{X}$ & 0.51 $\pm$ 0.02 & \textbf{0.77 $\pm$ 0.03} & 0.82 $\pm$ 0.02 & 0.82 $\pm$ 0.01 & 0.76 $\pm$ 0.01 \\
    \methodQ$_{Y}$ & 0.65 $\pm$ 0.02 & 0.76 $\pm$ 0.02 & \textbf{0.90 $\pm$ 0.01} & 0.91 $\pm$ 0.01 & 0.98 $\pm$ 0.01 \\
    \methodQ$_{X,Y}$ & 0.51 $\pm$ 0.02 & \textbf{0.77 $\pm$ 0.04} & 0.82 $\pm$ 0.10 & 0.83 $\pm$ 0.01 & 0.76 $\pm$ 0.01 \\
    \bottomrule
    \end{tabular}
    \label{tab:Q2app}
\end{table*}

\subsection{Training the rejector without augmenting the data}\label{app:uler-aug}
In this section, we evaluate whether augmentation improves the rejector’s performance, and thus whether collecting per-feature feedback is beneficial. To this end, we compare \method with an ablated variant, \method{\tt -NOAUG}, which does not leverage the feedback-aware augmentation strategy (\ie does not exploit the per-feature feedback). Specifically, \method{\tt -NOAUG} trains the rejector as described in \cref{sec:creating_dataset}, but uses $\calD$ instead of the augmented data $\calD_{aug}$.

\cref{tab:uler-aug} reports the average AUROC per dataset for \method and \method{\tt -NOAUG}, assessing their performance in distinguishing low-quality from high-quality explanations. For comparison, we also report the best-performing baseline from Q1 for each dataset, denoted as \textit{best}. \method consistently matches or outperforms its ablated variant across all considered datasets. While the overall performance gain is quite small ($\approx 2\%$), this improvement is highly reliable: \method never underperforms the variant without augmentation. We argue that this improvement is still worth it given the minimal additional cost to obtain the feature-level feedback. Once user-provided quality judgments are collected, obtaining per-feature feedback is inexpensive because users are already focused on identifying features with wrong scores to assess explanation quality. In cases where per-feature feedback is not available, one could skip the augmentation step and simply use \method{\tt -NOAUG}, which still consistently outperforms the \textit{best} baseline across all datasets and achieves an average AUROC that is at least $14\%$ higher.
\begin{table*}[!t]
  \caption{\textbf{\method shows a small but consistent improvement over its variant without augmentation in separating low-quality from high-quality explanations.}
  Average AUROC for \method and \method{\tt -NOAUG} across the five datasets. For comparison, we also report \textit{best}, the best performing baseline from Q2 for each dataset. \method consistently achieves a modest but consistent improvement in AUROC across all datasets, while \method{\tt -NOAUG} still always outperforms \textit{best}.}
    \setlength{\tabcolsep}{4pt}
    \fontsize{8pt}{8pt}\selectfont
    \centering
    \begin{tabular}{l|ccccc}
    \toprule
&  {\tt iknl} & {\tt sleepstage} & {\tt celeba} & {\tt adult} & {\tt compas}  \\
    \midrule
    \method & \textbf{0.67 $\pm$ 0.02} & \textbf{0.77 $\pm$ 0.02} & \textbf{0.90 $\pm$ 0.01} & \textbf{0.94 $\pm$ 0.01} & \textbf{0.99 $\pm$ 0.01} \\
    \midrule
    \method{\tt -NOAUG} & 0.63 $\pm$ 0.02 & 0.75 $\pm$ 0.03 & \textbf{0.90 $\pm$ 0.01} & 0.91 $\pm$ 0.01 & 0.96 $\pm$ 0.01 \\
\textit{best} & 0.62 $\pm$ 0.01 & 0.62 $\pm$ 0.02 & 0.75 $\pm$ 0.01 & 0.68 $\pm$ 0.02 & 0.75 $\pm$ 0.02 \\
    \bottomrule
    \end{tabular}
    \label{tab:uler-aug}
\end{table*}

\subsection{Robustness to the choice of the explainer} \
\label{app:lime_experiment}
In this section, we assess the robustness of our approach to the choice of explanation method. Specifically, we replicate the experimental setup from \cref{sec:Q1-Q2}, but generate all explanations using \LIME \cite{ribeiro2016should} with its default hyperparameters.

\begin{figure*}[ht]
\centering
    \includegraphics[width=\textwidth]{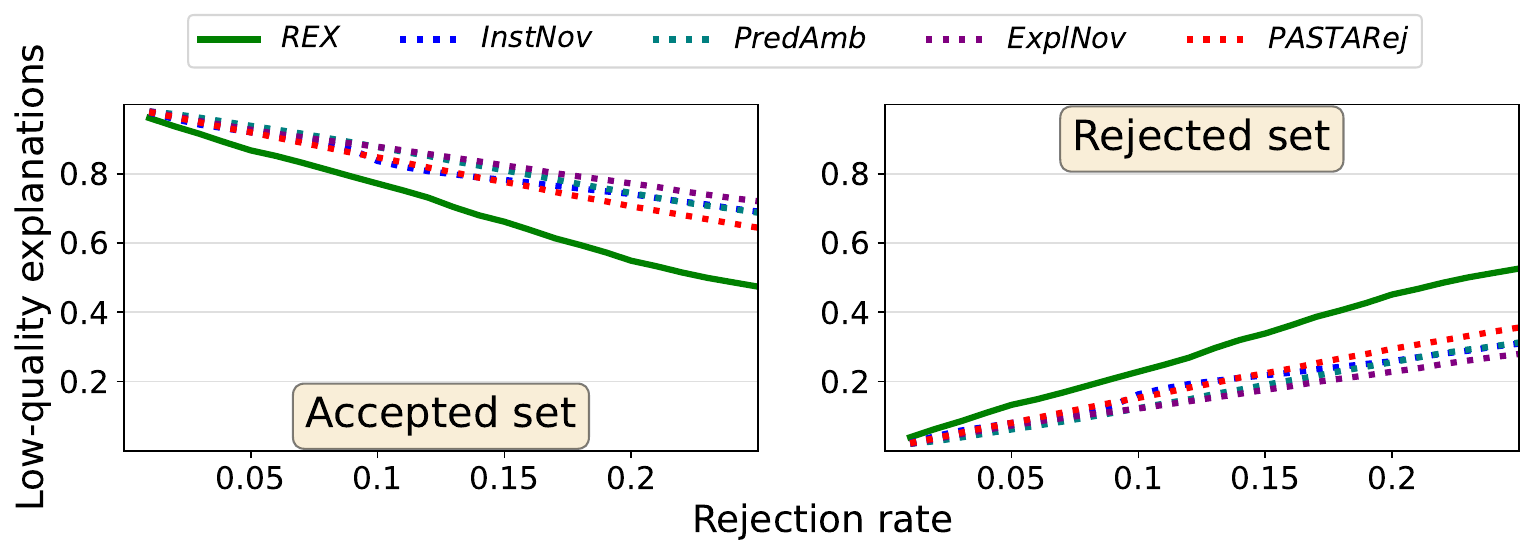}
  \caption{
  \textbf{\method rejects on average more low-quality explanations than all competitors when \LIME is used as explainer.}
  Average percentage of low quality explanations in the accepted and rejected set for all the considered strategies over the five datasets for $25$ rejection rates $\rho_\%$. \method outperforms all the competitors for all the considered rejection rates, demonstrating its robustness to the choice of the explainer.}
  \label{fig:lime}
  \vspace{4mm}
\end{figure*}

\textbf{Comparison with the competitors.} \cref{fig:lime} shows the percentage of low-quality explanations for the accepted and the rejected set as a function of the rejection rate $\rho_\%$ averaged across the five datasets considered. Even when using \LIME, \method outperforms the competitors across all rejection rates. On average, across all datasets and rejection rates, \method reduces the percentage of low-quality explanations in the accepted set by $10\%$ compared to the best competitor \PASTARej.
\\
\cref{tab:lime_q1} reports the average AUROC per dataset. Also in this case, \method consistently outperforms all baselines in all datasets by improving the AUROC by at least $18\%$ against the best competitor \PASTARej.

\textbf{\method against its ablated variants.} \cref{tab:lime_q2} reports the average AUROC per dataset for \method and its ablated variants that consider each metric individually. Across all datasets, \method more effectively distinguishes between high- and low-quality explanations than its ablated variants, achieving an average improvement of at least $10\%$ against the best competitor, \methodQ$_{plaus}$.
\begin{table*}[ht]
\caption{
  \textbf{\method outperforms the competitors at separating low-quality from high-quality explanations when \LIME is used as explainer.}
 Average AUROC ($\pm$ std) for all the rejection strategies over the five datasets. \method consistently obtains the best results in all datasets, demonstrating its robustness to the choice of the explainer}
    \setlength{\tabcolsep}{4pt}
    \fontsize{8pt}{8pt}\selectfont
    \centering
    \begin{tabular}{l|ccccc}
    \toprule
    & {\tt iknl} & {\tt sleepstage} & {\tt celeba} & {\tt adult} & {\tt compas} \\
    \midrule
    \method & \textbf{0.64 $\pm$ 0.01} & \textbf{0.99 $\pm$ 0.01} & \textbf{0.96 $\pm$ 0.01} & \textbf{0.98 $\pm$ 0.01} & \textbf{0.93 $\pm$ 0.01} \\
    \midrule
    \InstNov & 0.39 $\pm$ 0.01 & 0.54 $\pm$ 0.02 & 0.41 $\pm$ 0.01 & 0.44 $\pm$ 0.01 & 0.57 $\pm$ 0.02 \\
    \PredAmb & 0.47 $\pm$ 0.01 & 0.37 $\pm$ 0.02 & 0.60 $\pm$ 0.01 & 0.82 $\pm$ 0.01 & 0.56 $\pm$ 0.01 \\
    \ExplNov & 0.41 $\pm$ 0.01 & 0.44 $\pm$ 0.04 & 0.40 $\pm$ 0.02 & 0.61 $\pm$ 0.02 & 0.59 $\pm$ 0.02 \\
    \PASTARej & 0.53 $\pm$ 0.02 & 0.78 $\pm$ 0.10 & 0.85 $\pm$ 0.02 & 0.70 $\pm$ 0.11 & 0.68 $\pm$ 0.06 \\
\bottomrule
\end{tabular}
    \label{tab:lime_q1}
\end{table*}
\begin{table*}[ht]
\caption{
  \textbf{\method outperforms its variants at separating low-quality from high-quality explanations when \LIME is used as explainer.}
 Average AUROC ($\pm$ std) for \method and its ablated variants over the five datasets. }
    \setlength{\tabcolsep}{4pt}
    \fontsize{8pt}{8pt}\selectfont
    \centering
    \begin{tabular}{l|ccccc}
    \toprule
    & {\tt iknl} & {\tt sleepstage} & {\tt celeba} & {\tt adult} & {\tt compas} \\
    \midrule
    \method & \textbf{0.64 $\pm$ 0.01} & \textbf{0.99 $\pm$ 0.01} & \textbf{0.96 $\pm$ 0.01} & \textbf{0.98 $\pm$ 0.01} & \textbf{0.93 $\pm$ 0.01} \\
    \midrule
    \methodQ$_{faith}$ & 0.46 $\pm$ 0.01 & 0.58 $\pm$ 0.02 & 0.64 $\pm$ 0.01 & 0.92 $\pm$ 0.01 & 0.63 $\pm$ 0.01 \\
    \methodQ$_{stab}$ & 0.50 $\pm$ 0.01 & 0.55 $\pm$ 0.04 & 0.50 $\pm$ 0.01 & 0.50 $\pm$ 0.01 & 0.50 $\pm$ 0.01 \\
    \methodQ$_{plaus}$ & 0.60 $\pm$ 0.01 & 0.97 $\pm$ 0.01 & 0.91 $\pm$ 0.01 & 0.61 $\pm$ 0.10 & 0.90 $\pm$ 0.01 \\
\bottomrule
\end{tabular}
    \label{tab:lime_q2}
\end{table*}

\subsection{Simulating plausibility labels via LLM}\label{app:llm}
In this section, we repeat the experiment from Q1 and Q2, but we use a large language model (Llama-3.1-8B-Instruct) to simulate the plausibility labels $y^{plaus}$ and identify features with incorrect relevance scores. Following \cite{domnich2025towards}, we carefully crafted a prompt that (i) defines the evaluation task, (ii) introduces the structure and meaning of \SHAP explanations, and (iii) specifies the expected output format. The prompt used for the {\tt compas} dataset is provided below. This template can be easily adapted to other datasets by modifying the initial task description and the examples illustrating the \SHAP scores.

\begin{lstlisting}[language=Octave]
You are an expert in explainable AI and criminal justice risk assessment. Your task is to evaluate the quality of a SHAP explanation that describes why a person may be predicted to **recommit a crime**.

Each explanation is a list of features in the following format:
<featureID> <feature_name> : <feature_value> = <feature relevance score>

Your goal is to determine how **reasonable and high-quality** the explanation is, based on the SHAP scores and your domain knowledge.

### Understanding SHAP scores:
- A positive SHAP score (> 0) means the feature increases the risk of recidivism, contributing to a higher predicted risk.
- A negative SHAP score (< 0) means the feature decreases the risk, contributing to a lower predicted risk.
- A SHAP score of 0 means the feature has no impact on the prediction.
- The magnitude of the SHAP score reflects the strength of the feature's influence on the model's decision - larger absolute values imply greater impact.

### Your task:
Assign a **quality score from 1 to 5**:
- **5:** Excellent explanation - all important features have appropriate SHAP scores, and no suspicious or unjustified values.
- **4:** Good explanation - mostly reasonable, with at most minor issues in some features.
- **3:** Moderate quality - some questionable or poorly aligned SHAP scores, but overall still partially plausible.
- **2:** Poor quality - several features have inappropriate or suspicious SHAP scores.
- **1:** Very low quality - the explanation is clearly flawed, with major issues in multiple key features.

Also, list **the feature IDs** whose relevance scores are **unjustified or suspicious**, based on the feature's value and known importance.

Do not consider the model's prediction. Focus only on whether the explanation is plausible and grounded.

### Output format:
<score><space><comma-separated list of incorrect feature IDs>

Examples:
- Excellent explanation: `5`
- Good explanation with minor issues: `4 5`
- Low quality with clear issues: `2 1,6`
- Very low quality with major issues: `1 2,4,7`

If there are no suspicious features, leave the second part empty (just the score). DO NOT include any additional text or explanations in your response.
\end{lstlisting}

Based on the LLM's responses, the generated quality scores were converted into binary labels following the same procedure used in the user study (see \cref{sec:Q3}): explanations with scores strictly below three were classified as low-quality, while the others were deemed as high-quality.

\begin{figure*}[ht]
  \centering
  \includegraphics[width=1\textwidth]{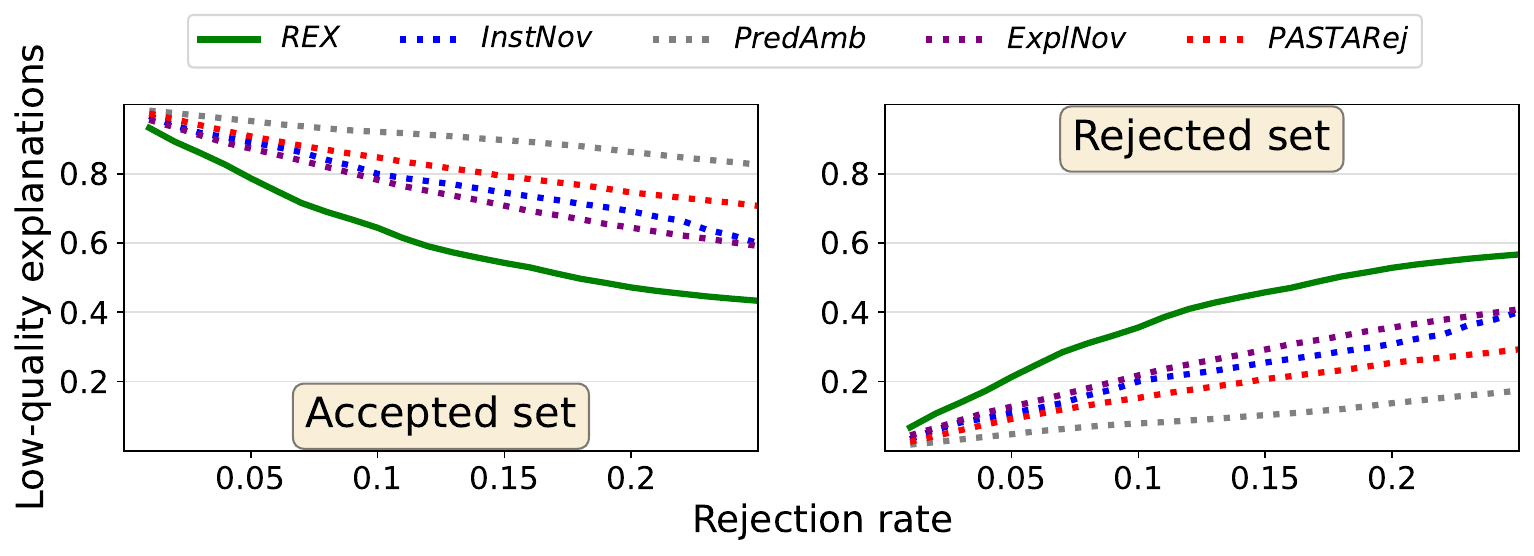}
  \caption{\textbf{\method rejects on average more low-quality explanations than all competitors when the simulated human-feedback is obtained via LLM.}
  Average percentage of low quality explanations in the accepted and rejected set for all the considered strategies over the five datasets for $25$ rejection rates $\rho_\%$.}
  \label{fig:llm}
  \end{figure*}

\textbf{Comparison with the competitors.}  
\cref{fig:llm} shows the percentage of low-quality explanations for the accepted and the rejected set as a function of the rejection rate $\rho_\%$ averaged over the five considered datasets. On average, \method rejects more low-quality explanations than the competitors: about $12\%$ more than \ExplNov, $17\%$ vs \InstNov, and over $20\%$ vs \PASTARej and \PredAmb. 

\cref{tab:llm_q1} reports the average AUROC per dataset. In all datasets, \method is better at distinguishing between high- and low-quality explanations than its competitors, with 
an average improvement of $13\%$ and $16\%$ over the best performing competitors \ExplNov and \InstNov.

\textbf{\method against its ablated variants.} \cref{tab:llm_q2} reports the average AUROC per dataset for \method and its ablated variants that consider each metric individually. Across all datasets, \method more effectively distinguishes between high- and low-quality explanations than its ablated variants, achieving an average improvement of at least $12\%$ against the best competitor, \methodQ$_{plaus}$. 
\begin{table*}[ht]
    \caption{
  \textbf{\method outperforms the competitors at separating low-quality  from high-quality explanations when the simulated human feedback is obtained via LLM.}
  Average AUROC ($\pm$ std) for all the rejection strategies over the five datasets. }
    \setlength{\tabcolsep}{4pt}
    \fontsize{8pt}{8pt}\selectfont
    \centering
    \begin{tabular}{l|ccccc}
    \toprule
&  {\tt iknl} & {\tt sleepstage} & {\tt celeba} & {\tt adult} & {\tt compas} \\
\midrule
\method & \textbf{0.72 $\pm$ 0.02} & \textbf{0.87 $\pm$ 0.02} & \textbf{0.79 $\pm$ 0.01} & \textbf{0.71 $\pm$ 0.01} & \textbf{0.70 $\pm$ 0.01} \\
\midrule
\PredAmb & 0.63 $\pm$ 0.02 & 0.27 $\pm$ 0.01 & 0.61 $\pm$ 0.01 & 0.52 $\pm$ 0.01 & 0.52 $\pm$ 0.01 \\
\InstNov & 0.63 $\pm$ 0.02 & 0.79 $\pm$ 0.02 & 0.52 $\pm$ 0.02 & 0.54 $\pm$ 0.01 & 0.46 $\pm$ 0.01 \\
\ExplNov & 0.63 $\pm$ 0.03 & 0.82 $\pm$ 0.01 & 0.69 $\pm$ 0.01 & 0.56 $\pm$ 0.01 & 0.48 $\pm$ 0.01 \\
    \PASTARej & 0.52 $\pm$ 0.10 & 0.56 $\pm$ 0.08 & 0.49 $\pm$ 0.03 & 0.48 $\pm$ 0.03 & 0.54 $\pm$ 0.03\\
\bottomrule
\end{tabular}
    \label{tab:llm_q1}
\end{table*}
\begin{table*}[ht]
\caption{
  \textbf{\method outperforms its variants at separating low-quality from high-quality explanations when the simulated human feedback is obtained via LLM.}
 Average AUROC ($\pm$ std) for \method and its ablated variants over the five datasets.}
    \setlength{\tabcolsep}{4pt}
    \fontsize{8pt}{8pt}\selectfont
    \centering
    \begin{tabular}{l|ccccc}
    \toprule
    & {\tt iknl} & {\tt sleepstage} & {\tt celeba} & {\tt adult} & {\tt compas} \\
    \midrule
\method & \textbf{0.72 $\pm$ 0.02} & \textbf{0.87 $\pm$ 0.02} & \textbf{0.79 $\pm$ 0.01} & \textbf{0.71 $\pm$ 0.01} & \textbf{0.70 $\pm$ 0.01} \\
    \midrule
    \methodQ$_{faith}$ & 0.71 $\pm$ 0.02 & 0.70 $\pm$ 0.03 & 0.53 $\pm$ 0.01 & 0.55 $\pm$ 0.02 & 0.63 $\pm$ 0.01 \\
    \methodQ$_{stab}$ & 0.52 $\pm$ 0.01 & 0.85 $\pm$ 0.02 & 0.72 $\pm$ 0.01 & 0.59 $\pm$ 0.01 & 0.62 $\pm$ 0.01 \\
    \methodQ$_{human}$ & 0.53 $\pm$ 0.06 & 0.83 $\pm$ 0.04 & 0.59 $\pm$ 0.02 & 0.51 $\pm$ 0.02 & 0.58 $\pm$ 0.01 \\
\bottomrule
\end{tabular}
    \label{tab:llm_q2}
\end{table*}

\subsection{Experimenting with different thresholds for the stability and faithfulness labels}\label{app:different_thresholds}
In this section, we repeat the experiments from Q1 and Q2 to demonstrate that our findings are robust to the exact threshold used to binarize the stability and faithfulness metrics. Specifically, we evaluate two alternative thresholds $u_{\%}=5\%$ and $u_{\%}=20\%$ to obtain $y^{stab}$ and $y^{faith}$.

\begin{figure*}[ht]
  \centering
  \includegraphics[width=1\textwidth]{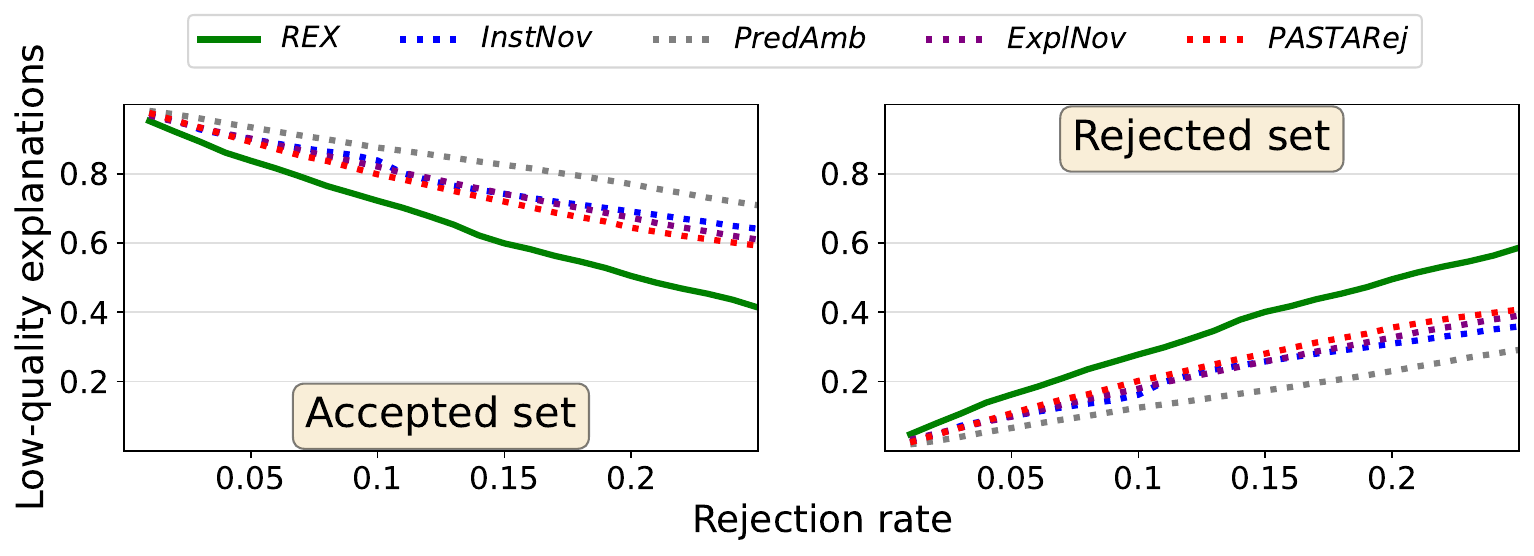}
  \caption{\textbf{\method rejects on average more low-quality explanations than all competitors when $u_{\%} = 5\%$.}
  Average percentage of low quality explanations in the accepted and rejected set for all the considered strategies over the five datasets for $25$ rejection rates $\rho_\%$.}
  \label{fig:threshold05}
  \end{figure*}

\begin{figure*}[ht]
  \centering
  \includegraphics[width=1\textwidth]{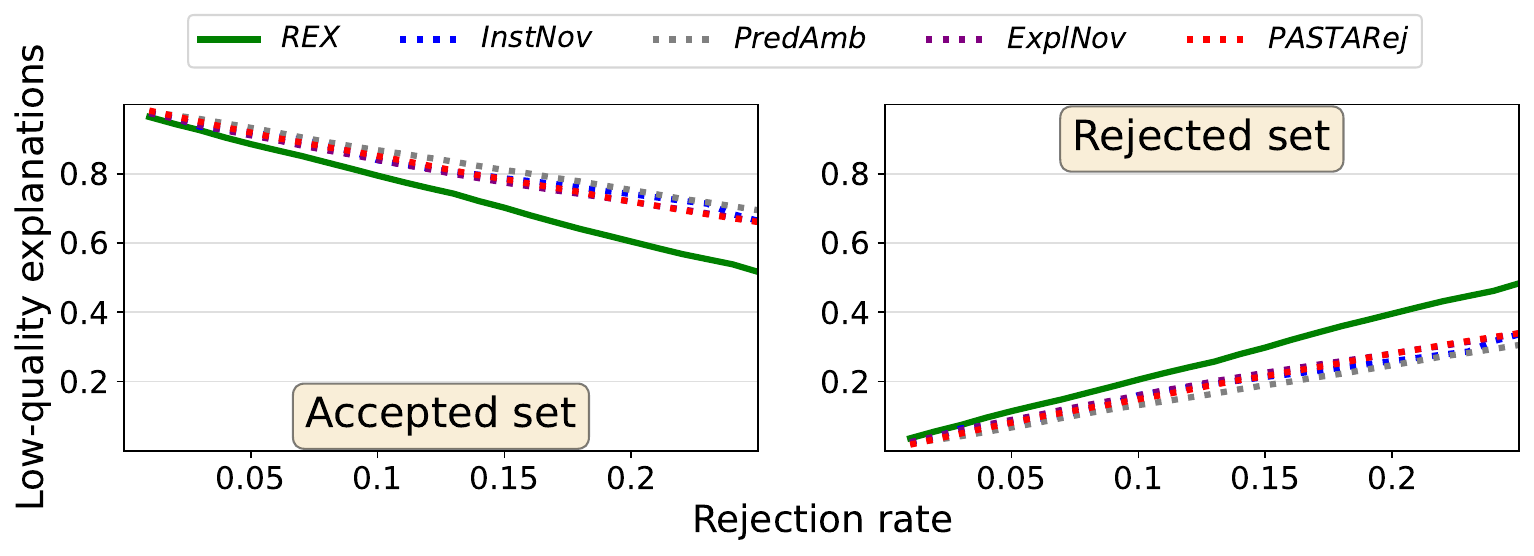}
  \caption{\textbf{\method rejects on average more low-quality explanations than all competitors when $u_{\%} = 20\%$.}
  Average percentage of low quality explanations in the accepted and rejected set for all the considered strategies over the five datasets for $25$ rejection rates $\rho_\%$.}
  \label{fig:threshold20}
  \end{figure*}

\textbf{Comparison with the competitors.}  
\cref{fig:threshold05} and \cref{fig:threshold20} show the percentage of low-quality explanations for the accepted and the rejected set as a function of the rejection rate $\rho_\%$ averaged over the five considered datasets for $u_{\%} = 5\%$ and $u_{\%} = 20\%$, respectively. For both thresholds, \method consistently rejects the highest number of low-quality explanations

\cref{tab:threshold05_q1} and \cref{tab:threshold20_q1} report the average AUROC per dataset for $u_{\%} = 5\%$ and $u_{\%} = 20\%$, respectively. Across all datasets, \method remains better at distinguishing between high- and low-quality explanations than its competitors, confirming its effectiveness irrespective of the chosen threshold.

\textbf{\method against its ablated variants.} \cref{tab:threshold05_q2} and \cref{tab:threshold20_q2} report the average AUROC per dataset for \method and its ablated variants that consider each metric individually when $u_{\%} = 5\%$ and $u_{\%} = 20\%$, respectively. \method more effectively distinguishes between high- and low-quality explanations than its ablated variants, achieving an average improvement of at least $8\%$ over the strongest single-metric competitor \methodQ$_{plaus}$. 
\begin{table*}[ht]
    \caption{
  \textbf{\method outperforms the competitors at separating low-quality from high-quality explanations when $u_{\%} = 5\%$.}
  Average AUROC ($\pm$ std) for all the rejection strategies over the five datasets. }
    \setlength{\tabcolsep}{4pt}
    \fontsize{8pt}{8pt}\selectfont
    \centering
    \begin{tabular}{l|ccccc}
    \toprule
&  {\tt iknl} & {\tt sleepstage} & {\tt celeba} & {\tt adult} & {\tt compas} \\
\midrule
\method & \textbf{0.61 $\pm$ 0.02} & \textbf{0.73 $\pm$ 0.02} & \textbf{0.87 $\pm$ 0.01} & \textbf{0.90 $\pm$ 0.01} & \textbf{0.98 $\pm$ 0.01} \\
\midrule
\PredAmb & 0.59 $\pm$ 0.02 & 0.28 $\pm$ 0.02 & 0.65 $\pm$ 0.01 & 0.59 $\pm$ 0.01 & 0.54 $\pm$ 0.02 \\
\InstNov & 0.59 $\pm$ 0.02 & 0.53 $\pm$ 0.03 & 0.47 $\pm$ 0.01 & 0.69 $\pm$ 0.01 & 0.76 $\pm$ 0.01 \\
\ExplNov & 0.56 $\pm$ 0.01 & 0.52 $\pm$ 0.03 & 0.68 $\pm$ 0.01 & 0.69 $\pm$ 0.02 & 0.78 $\pm$ 0.01 \\
\PASTARej & 0.55 $\pm$ 0.06 & 0.62 $\pm$ 0.06 & 0.65 $\pm$ 0.02 & 0.59 $\pm$ 0.08 & 0.81 $\pm$ 0.01 \\
\bottomrule
\end{tabular}
    \label{tab:threshold05_q1}
\end{table*}
\begin{table*}[ht]
\caption{
  \textbf{\method outperforms its variants at separating low-quality from high-quality explanations when $u_{\%}=5\%$}
 Average AUROC ($\pm$ std) for \method and its ablated variants over the five datasets.}
    \setlength{\tabcolsep}{4pt}
    \fontsize{8pt}{8pt}\selectfont
    \centering
    \begin{tabular}{l|ccccc}
    \toprule
    & {\tt iknl} & {\tt sleepstage} & {\tt celeba} & {\tt adult} & {\tt compas} \\
    \midrule
\method & \textbf{0.61 $\pm$ 0.02} & 0.73 $\pm$ 0.02 & \textbf{0.87 $\pm$ 0.01} & \textbf{0.90 $\pm$ 0.01} & \textbf{0.98 $\pm$ 0.01} \\
\midrule
\methodQ$_{faith}$ & 0.60 $\pm$ 0.01 & 0.65 $\pm$ 0.02 & 0.58 $\pm$ 0.01 & 0.62 $\pm$ 0.02 & 0.54 $\pm$ 0.02 \\
\methodQ$_{faith}$ & 0.52 $\pm$ 0.01 & 0.50 $\pm$ 0.01 & 0.70 $\pm$ 0.01 & 0.67 $\pm$ 0.01 & 0.40 $\pm$ 0.01 \\
\methodQ$_{plaus}$ & 0.55 $\pm$ 0.05 & \textbf{0.77 $\pm$ 0.02} & 0.78 $\pm$ 0.02 & 0.73 $\pm$ 0.05 & 0.88 $\pm$ 0.01 \\
\bottomrule
\end{tabular}
    \label{tab:threshold05_q2}
\end{table*}
\begin{table*}[ht]
    \caption{
  \textbf{\method outperforms the competitors at separating low-quality from high-quality explanations when $u_{\%} = 20\%$.}
  Average AUROC ($\pm$ std) for all the rejection strategies over the five datasets. }
    \setlength{\tabcolsep}{4pt}
    \fontsize{8pt}{8pt}\selectfont
    \centering
    \begin{tabular}{l|ccccc}
    \toprule
&  {\tt iknl} & {\tt sleepstage} & {\tt celeba} & {\tt adult} & {\tt compas} \\
\midrule
\method & \textbf{0.75 $\pm$ 0.01} & \textbf{0.83 $\pm$ 0.02} & \textbf{0.95 $\pm$ 0.01} & \textbf{0.97 $\pm$ 0.01} & \textbf{0.99 $\pm$ 0.01} \\
\midrule
\PredAmb & 0.68 $\pm$ 0.01 & 0.22 $\pm$ 0.02 & 0.76 $\pm$ 0.01 & 0.58 $\pm$ 0.02 & 0.59 $\pm$ 0.01 \\
\InstNov & 0.67 $\pm$ 0.01 & 0.55 $\pm$ 0.03 & 0.50 $\pm$ 0.02 & 0.57 $\pm$ 0.01 & 0.64 $\pm$ 0.02 \\
\ExplNov & 0.60 $\pm$ 0.02 & 0.59 $\pm$ 0.03 & 0.82 $\pm$ 0.01 & 0.69 $\pm$ 0.01 & 0.62 $\pm$ 0.01 \\
\PASTARej & 0.64 $\pm$ 0.10 & 0.62 $\pm$ 0.07 & 0.67 $\pm$ 0.04 & 0.49 $\pm$ 0.08 & 0.69 $\pm$ 0.02 \\
\bottomrule
\end{tabular}
    \label{tab:threshold20_q1}
\end{table*}
\begin{table*}[ht]
\caption{
  \textbf{\method outperforms its variants at separating low-quality from high-quality explanations when $u_{\%} = 20\%$.}
 Average AUROC ($\pm$ std) for \method and its ablated variants over the five datasets.}
    \setlength{\tabcolsep}{4pt}
    \fontsize{8pt}{8pt}\selectfont
    \centering
    \begin{tabular}{l|ccccc}
    \toprule
    & {\tt iknl} & {\tt sleepstage} & {\tt celeba} & {\tt adult} & {\tt compas} \\
    \midrule
\method & \textbf{0.75 $\pm$ 0.01} & \textbf{0.83 $\pm$ 0.02} & \textbf{0.95 $\pm$ 0.01} & \textbf{0.97 $\pm$ 0.01} & \textbf{0.99 $\pm$ 0.01} \\
    \midrule
\methodQ$_{faith}$ & \textbf{0.75 $\pm$ 0.01} & 0.74 $\pm$ 0.02 & 0.62 $\pm$ 0.01 & 0.59 $\pm$ 0.02 & 0.69 $\pm$ 0.01 \\
\methodQ$_{stab}$ & 0.50 $\pm$ 0.01 & 0.61 $\pm$ 0.01 & 0.82 $\pm$ 0.01 & 0.74 $\pm$ 0.01 & 0.59 $\pm$ 0.01 \\
\methodQ$_{plaus}$ & 0.65 $\pm$ 0.02 & 0.76 $\pm$ 0.02 & 0.79 $\pm$ 0.02 & 0.61 $\pm$ 0.05 & 0.81 $\pm$ 0.01 \\
\bottomrule
\end{tabular}
    \label{tab:threshold20_q2}
\end{table*}

\subsection{User study: confidence interval plot}\label{app:Q1ci}
\begin{figure*}[h]
\centering
    \includegraphics[width=\textwidth]{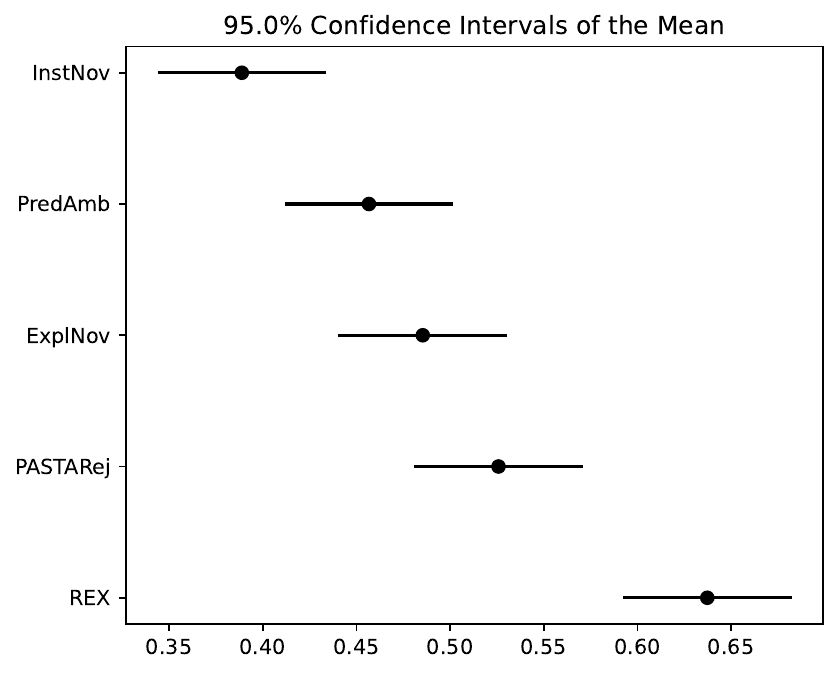}
  \caption{\textbf{Statistical significance analysis of AUROC performance.} $95\%$ confidence intervals for \method and competitors using Tukey’s HSD test. Results are aggregated over 10 independent runs. Non-overlapping intervals indicate statistically significant differences at $\alpha = 0.05$.}
  \label{fig:Q1ci}
\end{figure*}

\cref{fig:Q1ci} reports the $95\%$ confidence interval plot for the statistical analysis run in Q3. \method statistically outperforms all the competitors.

\section{LLM usage} LLMs were used to polish the writing, to rephrase sentences, and to debug the code. Our manuscript and our code was first human-generated, and then possibly enhanced by LLMs. 

\begin{figure}
    \centering
    \includegraphics[width=0.3\linewidth]{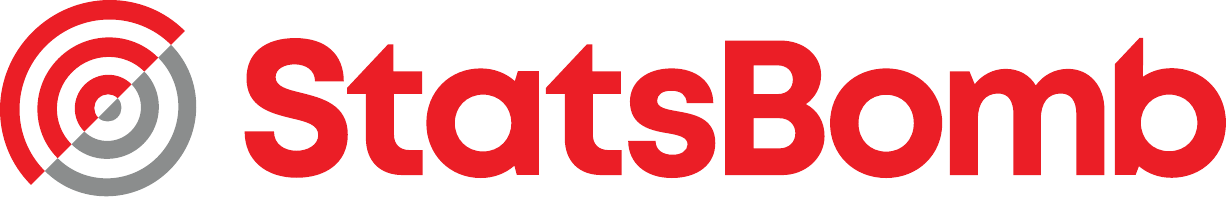}
\end{figure}